\def\year{2021}\relax
\documentclass[letterpaper]{article} % DO NOT CHANGE THIS
\usepackage{aaai21}  % DO NOT CHANGE THIS
\usepackage{times}  % DO NOT CHANGE THIS
\usepackage{helvet} % DO NOT CHANGE THIS
\usepackage{courier}  % DO NOT CHANGE THIS
\usepackage[hyphens]{url}  % DO NOT CHANGE THIS
\usepackage{graphicx} % DO NOT CHANGE THIS
\usepackage{natbib}  % DO NOT CHANGE THIS AND DO NOT ADD ANY OPTIONS TO IT
\usepackage{caption} % DO NOT CHANGE THIS AND DO NOT ADD ANY OPTIONS TO IT
\usepackage{amsfonts}       % blackboard math symbols
\usepackage{amsmath}
\usepackage{multirow}
\usepackage{bm}
\usepackage{color}
\usepackage{amsthm}
\usepackage{exscale}
\usepackage{relsize}
\usepackage{xr}
\newcommand{\comment}[1]{}

\newcommand{\et}{\emph{et al.}}

\newcommand{\eg}{\emph{e.g.}}
\usepackage[utf8]{inputenc} % allow utf-8 input
\usepackage{url}            % simple URL typesetting
\usepackage{booktabs}       % professional-quality tables
\usepackage{multirow}
\usepackage{amsfonts}       % blackboard math symbols
\usepackage{nicefrac}       % compact symbols for 1/2, etc.
\usepackage{microtype}      % microtypography
\usepackage{algpseudocode} 
\usepackage{algorithm}
\usepackage{amsmath}
\usepackage{graphicx}
\usepackage{graphics}
\usepackage{wrapfig}
\usepackage[switch]{lineno}
  
\title{Adversarial Meta Sampling for Multilingual Low-Resource Speech Recognition}

\author{

    Yubei Xiao\textsuperscript{\rm 1}, Ke Gong\textsuperscript{\rm 3}, Pan Zhou\textsuperscript{\rm 4}, Guolin Zheng\textsuperscript{\rm 1}, Xiaodan Liang\textsuperscript{\rm 2,3}, Liang Lin\textsuperscript{\rm 1,3}\thanks{Liang Lin is the corresponding author of this work.}
}
\affiliations{
    \textsuperscript{\rm 1} School of Computer Science and Engineering, Sun Yat-sen University, China \\
    \textsuperscript{\rm 2} School of Intelligent Systems Engineering, Sun Yat-sen University, China \\
    \textsuperscript{\rm 3} Dark Matter AI Research,
    \textsuperscript{\rm 4} SalesForce
    \{xiaoyb5,zhengglin\}@mail2.sysu.edu.cn, \{kegong936,xdliang328\}@gmail.com, pzhou@salesforce.com, linliang@ieee.org

}

\begin{document}
% \linenumbers
\maketitle

\begin{abstract}
Low-resource automatic speech recognition (ASR) is challenging, as the low-resource target language data cannot well train an ASR model. To solve this issue, meta-learning formulates ASR for each source language into many small ASR tasks and meta-learns a model initialization on all tasks from different source languages to access fast adaptation on unseen target languages.  However,  for different source languages, the quantity and difficulty vary greatly because of their different data scales and diverse phonological systems, which leads to task-quantity and task-difficulty imbalance issues and thus a failure of multilingual meta-learning ASR (MML-ASR).   In this work, we solve this problem by developing a novel adversarial meta sampling (AMS) approach to improve MML-ASR. When sampling tasks in MML-ASR, AMS adaptively determines the task sampling probability for each source language. Specifically, for each source language, if the query loss is large, it means that its tasks are not well sampled to train ASR model in terms of its quantity and difficulty and thus should be sampled more frequently for extra learning.  Inspired by this fact, we feed the historical task query loss of all source language domain into a  network to learn a task sampling policy for adversarially increasing the current query loss of MML-ASR. Thus, the learnt task sampling policy can master the learning situation of each language and thus predicts good task sampling probability for each language for more effective learning.  Finally, experiment results on two multilingual datasets show significant performance improvement when applying our AMS on MML-ASR, and also demonstrate the applicability of AMS to other low-resource speech tasks and transfer learning ASR approaches.
% Our codes are available at: \href{https://github.com/iamxiaoyubei/AMS}{https://github.com/iamxiaoyubei/AMS}.
%  All source codes, models and datasets have been attached in the supplementary.
\end{abstract}

% \vspace{-0.6em}
\section{Introduction}\label{introduction}
Automatic Speech Recognition (ASR) has attracted a lot of attention recently  and achieved significant improvements~\cite{Chan2016ListenAA,Graves2006ConnectionistTC,Pratap2019Wav2LetterAF} brought by the success of deep neural networks. However, building an end-to-end deep ASR model  often requires huge transcribed training data, which is impractical for the low-resource languages due to the scarcity of audio data and the huge labor resources consumed in transcription.
 
\begin{figure*}
	\centering
	\includegraphics[width=0.8\linewidth]{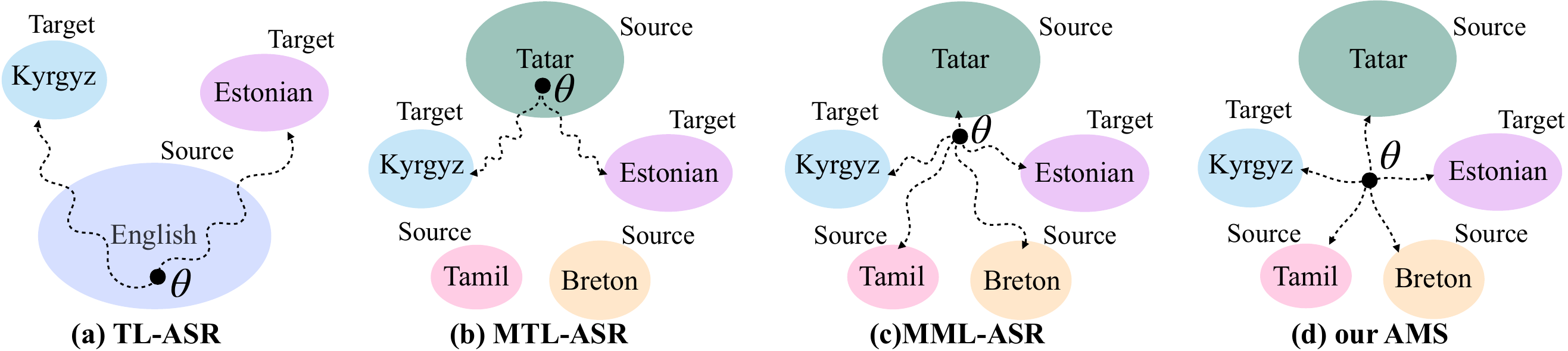}
% 	\vspace{-3mm}
% 	\vspace{-6mm}
	\caption{Comparison of learnt initializations under task-quantity-balanced sampling.  The dashed lines denote the adaptation paths from  initialization  $\theta$ to target languages and the circular area of the language represents the training data scale of the language. (a) Initialization learnt in TL-ASR overfits  the only one source language. Initializations in MTL-ASR  (b) and MML-ASR   (c) are close to the optimal model of the large-scaled  language and  departure from the small-scaled languages.   (d) Initialization learnt by our AMS has a more balanced distance to all languages because of our adaptive sampling to handle the task imbalance. }
	%	\vspace{-8mm}
	\label{fig:motivation}
\end{figure*}

To solve this issue, many works are devoted to develop low-resource ASR approaches.   The  representative methods in this line are  transfer learning ASR (\textbf{TL-ASR})~\cite{Hu2019,Kunze2017TransferLF}, multilingual transfer learning ASR (\textbf{MTL-ASR})~\cite{Adams2019MassivelyMA,Cho2018MultilingualSS,Tong2017MultilingualTA} and multilingual meta-learning ASR (\textbf{MML-ASR})~\cite{Hsu2019MetaLF}  that all aim to learn an ASR model initialization  from source languages such that the  initialization can quickly adapt to target language via fine-tuning on a few data.  Among them, TL-ASR considers  one source language and  regards the pretrained ASR model on the source data as a model initialization. But as shown in Fig.~\ref{fig:motivation} (a), the learnt initialization by TL-ASR  often overfits 
the source language and cannot quickly adapt  to a different target language. To resolve this issue,  MTL-ASR and MML-ASR consider multiple source languages. Inspired by multi-task learning, they both sample partial data from each language domain to construct a small speech recognition \textbf{task}.  Then for each sampled task, MTL-ASR  directly trains its model on this  task, while  MML-ASR adapts  its ASR model  to the validation data of the task via fine-tuning on a few training data of the task and minimizes the validation loss.  In this way, the learnt initializations by MTL-ASR and MML-ASR  can usually fast adapt to the target low-resource language, as both MTL-ASR and MML-ASR  learn the common knowledge from all tasks from different language domains which  facilitates learning target languages.

However, when  sampling tasks from these language domains, MTL-ASR and MML-ASR often ignore the underlying task imbalance issues which could result in unsatisfactory performance. First, different kinds of languages have different training data scales so the underlying task quantity for each language domain varies greatly, which leads to the \textbf{task-quantity imbalance}. Second, as different languages have diverse phonological systems, the tasks drawn from different language domains have various recognition difficulties, causing the \textbf{task-difficulty imbalance}. In this way, for MTL-ASR and MML-ASR, both \textbf{uniform sampling} that uniformly samples tasks from each language domain and  more advanced \textbf{task-quantity-balanced sampling} whose sampling rate for each domain positively relies on its task quantity (data scale) cannot handle the task imbalance issue.  Uniform sampling neither considers the task-quantity imbalance nor the task-difficulty imbalance, while task-quantity-balanced sampling directly ignores task-difficulty imbalance. So the learnt initializations by MTL-ASR and MML-ASR are biased, and cannot fast adapt to the target language. For instance,  as shown in Fig.~\ref{fig:motivation} (b) and (c),  when using task-quantity-balanced sampling in MTL-ASR and MML-ASR, their learnt initializations are often close to the optimal model of the large-scaled  language and departure from those of the small-scaled languages. So the learnt initializations are often far from the optimal models of target languages which usually locates around all source languages, and  cannot be fast adapted to target languages via fine-tuning on low-resource training data.

\textbf{Contributions.} To resolve the above task imbalance issue,  we develop a novel adversarial meta sampling  (\textbf{AMS}) method for multilingual low-resource speech recognition. This AMS can effectively help both MML-ASR and MLT-ASR handle task imbalance problem and boost their performance.   
%To overcome the language imbalance problem, our Adversarial Meta Sampling mechanism can improve MML-ASR models by adaptively sampling tasks in meta-training process to balance the difficulty and quantity of different language tasks. 
Considering the superior performance of MML-ASR over MLT-ASR, in this work we spend more efforts to introduce AMS on MML-ASR.  Specifically, we observe that the query losses of tasks from each language domain can well measure both task-quantity imbalance and task-difficulty imbalance. 
% Since if tasks from one language domain are not well sampled in terms of its task quantity and task difficulty, then they will have relatively large query losses. 
For each language domain, if its tasks are not well sampled in terms of its task quantity and task difficulty, then its tasks will have relatively large query losses.
It means that the language domain with large task query loss requires more training. So we design a policy network to increase the query loss of MML-ASR model through adversarial learning for sampling from proper language domain. Our policy network incorporates LSTM~\cite{Hochreiter1997LongSM} structure and attention mechanism to adaptively predict the most befitting task sampling probability for each language domain by using the long-term information in LSTM and the current query losses at each training iteration. Through such an online and adversarial manner, the sampling policy is dynamically changed along with the training state of the MML-ASR. In this way, the language domain that are not well sampled in terms of its task quantity and task difficulty will be sampled more in the later training iterations, making the MML-ASR model learn a more balanced initialization for better adaptation to the target languages as shown in Fig.~\ref{fig:motivation} (d).

% \pz{Specifically, we  design a policy network to sample the currently most benefiting language tasks in each meta-training step to enhance the meta-learning under the task-imbalance conditions, which can be optimized with the MML-ASR model in an adversarial and online manner. The sample policy is dynamically changed along with the training state, rather than fixed as a random strategy throughout the whole training like traditional meta-learning methods~\cite{Finn2017ModelAgnosticMF}. Under this strategy, the policy network can utilize query losses of the current step to generate more difficult language tasks learned insufficiently, which finally makes the ASR model learn a more balance initialization for better adaptation to the target languages, as shown in Fig.~\ref{fig:motivation} (d). Furthermore, our AMS method can be easy generalized and incorporated into Multilingual Transfer Learning ASR networks, to alleviate their overfitting problem and improve their performances. Please follow Sec. 4.1 to introduce it.}

% Additionally, the query loss of meta-training is used to guide the policy network to generate more difficult tasks learned insufficiently, which makes the model learn a more balanced initialization for better adaptation, as shown in Fig~\ref{fig:motivation}.

Moreover, we validate our method on several datasets with diverse languages selected from Mozilla Common Voice Corpus~\cite{commonV} and the public IARPA BABEL dataset~\cite{Gales2014SpeechRA}. The experimental results demonstrate that our AMS significantly improves the performance over the existing approaches on low-resource ASR, especially under the realistic task-imbalance scenarios. Furthermore, we conduct experiments on speech classification and speech translation, which proves that our AMS can be easily generalized to improve other low-resource speech tasks.

%Our contributions can be summarized as follows: 1) We investigate a challenging multilingual low-resource ASR task under a realistic scenarios where the languages and the data sizes differ largely. To alleviate the severe initialization problems of existing methods under the task-imbalance situation, we introduce an  Adversarial Meta Sampling framework that can adaptively sample tasks to learning a better model initialization for target low-resource languages. 2) We conduct extensive experiments with various languages on low-resource ASR task, which shows that our method significantly outperforms the existing low-resource ASR methods and achieves the state-of-the-art results. 3) Our method can also be easily extended to multilingual transfer learning and other speech tasks under low-resource setting.  The experiment results demonstrate the generalization ability of our method which largely improve the performance of speech classification and speech translation.

\section{Related Work}
\noindent\textbf{Transfer learning  ASR.} To alleviate the need for labeled data, recent works utilize unsupervised pre-training and semi-supervised methods to exploit unlabeled data, \eg~wav2vec~\cite{Schneider2019wav2vecUP}, predictive coding~\cite{Chung2019GenerativePF}, self-training~\cite{Kahn2019SelfTrainingFE} and weak distillation~\cite{Li2019SemisupervisedTF}. But they still require substantial unlabeled data which is unavailable for some minority languages.
% But   they still require  substantial  unlabeled data which is  unavailable for the practical   low-resource settings.  
 To solve this issue, 
 transfer learning is explored  via using other source languages to improve the performance of low-resource languages~\cite{Kunze2017TransferLF}, which requires that the  source and target  languages are similar and the source language has sufficiently large data. Moreover, multilingual transfer learning ASR~\cite{Dalmia2018SequenceBasedML,Watanabe2017LanguageIE,Toshniwal2018MultilingualSR} is developed using different languages to learn language-independent  representations for performance improvement under the low-resource setting. 
% The dominant architecture has been the so-called "multilingual bottleneck" model with shared hidden layers and many language-specific heads~\cite{Dalmia2018SequenceBasedML,Grzl2014AdaptationOM,Fr2015MultilingualBF}. Alternatively, some approaches constructed a shared language-independent phone set for each languages~\cite{Vu2014MultilingualDN,Tong2017AnIO} and some build an end-to-end ASR model for all the languages~\cite{Watanabe2017LanguageIE,Toshniwal2018MultilingualSR}. 
% But all of these transfer learning methods often overfit source languages and cannot well generalize. 
% But transfer learning methods are easy to overfit source languages and cannot well generalize.
 %Therefore, meta-learning was introduced to low-resource ASR to learn better initialization parameters for fast adaptation.\xyb{m} 

\noindent\textbf{Meta-learning ASR.}  
% Some  works introduced meta-learning~\cite{Finn2017ModelAgnosticMF,Nichol2018OnFM}  into low-resource natural language processing, such as NLU~\cite{Dou2019InvestigatingMA}, NMT~\cite{Gu2018MetaLearningFL}, and NLG~\cite{Mi2019MetaLearningFL}. This is because meta-learning approaches  can meta-learn a model initialization from training tasks with fast adaptation ability to new tasks with only a few data and thus is suitable to handle low-resource data learning problems. 
Meta-learning approaches~\cite{Zhou2019EfficientML,Zhou2020MAML} can meta-learn a model initialization from training tasks with fast adaptation ability to new tasks with only a few data and thus is suitable to handle low-resource data learning problems. 
Especially, Hsu~\et (Hsu et al. 2020) and Winata~\et~\cite{Winata2020MetaTransferLF} adopted MAML (Finn et al. 2017) for low-resource ASR and code-switched ASR and both achieved promising results.
% ~\cite{Hsu2019MetaLF}
%MAML  ~\cite{Finn2017ModelAgnosticMF}
% and achieved promising results. 
% But as mentioned in Sec.~\ref{introduction}, 
But these method ignores task imbalance in real-world scenarios and equally utilizes the meta-knowledge across all the languages, which leads to performance degradation.  %Furthermore, Bayesian TAML~\cite{lee2020l2b} was proposed to learn variables to balance the effect of meta- and task-specific learning under imbalance setting in image classification.
To alleviate quantity imbalance, Wang~\et~\cite{wang-etal-2020-balancing} improves differentiable data selection by optimizing a scorer with the average loss from different languages to balance the usage of data in multilingual model training. Besides the language quantity, our AMS also considers the language difﬁculty and learns the sampling policy in an adversarial manner.
% To solve this  task imbalance issue, we develop a task sampling method considering both task quantity and task difficulty and its effectiveness is verified by extensive experimental results. 
% \cite{wang-etal-2020-balancing}

\noindent\textbf{Adversarial learning ASR.}
Inspired by domain adversarial training~\cite{Ganin2016DomainAdversarialTO}, recent works introduced adversarial learning into ASR to learn robust features  invariant to noise conditions~\cite{Shinohara2016AdversarialML} and accents~\cite{Sun2018DomainAT}. Besides, some researchers use a domain-adversarial classification objective over many languages on multilingual ASR framework to force the shared layers to learn language-independent representations~\cite{Yi2018AdversarialMT}.
% learn representations taht are invariant to language.
% Moreover, domain adversarial training was applied into multilingual bottleneck networks to force the shared layers to learn language-invariant features~\cite{Yi2018AdversarialMT,Adams2019MassivelyMA}~\pz{(1)not clear: invariant to what, (2) this sentence structure is not good}.  
%There were also some works that combine adversarial learning with meta-learning~\cite{Yin2018AdversarialM,Zugner2019AdversarialAO,Goldblum2019RobustFL,Goodfellow2014ExplainingAH} to improve the robustness of model in few-shot tasks. 
%Moreover, adversarial training is also widely used in data augmentation~\cite{Cubuk2018AutoAugmentLA,Peng2018JointlyOD,Zhang2019AdversarialA,Luo2020LearnTA}, which tried to design an augmentation network that competes against a target network by generating hard augmentation operations.
Differently, our proposed method explores adversarial learning to solve the task imbalance problem in multilingual meta-learning ASR and can learn to adaptively sample the meta-training tasks  for effectively training low-resource ASR models.

\section{Preliminaries}
Here we briefly introduce the ASR model and its meta-learning version which are used in our method.
% \vspace{-0.8em}
\subsection{Multilingual Speech Recognition}
%\subsubsection{ASR model}
\noindent\textbf{ASR model.} 
%Our AMS framework can be implemented via  any end-to-end ASR networks. 
We first introduce the joint CTC-attention based end-to-end ASR architecture~\cite{Kim2017JointCB,Hori2017AdvancesIJ} because of its effectiveness and efficiency. It consists of a Seq2Seq network for frames alignment and symbols recognition, and a connectionist temporal classification (CTC) module~\cite{Graves2006ConnectionistTC} to encourage the alignments to be monotonic. 
For the seq2seq model, it  contains  an encoder, a decoder, and an attention unit. 
% Among them, the encoder extracts  hidden representation $H\!=\!\{h_i\}_{i=1}^L$ of the utterance $X\!=\!\{x_i\}_{i=1}^T\ (L\!\leq \!T)$. The attention unit takes  the previous decoder state $s_{i-1}$ and representation $H$ as input, and then generates the context vector $c_{i}$. When fed  the previous decoded output $y_{i-1}$ and the  context vector $c_{i}$, the decoder  can  predict the output $y_{i}$. Then the seq2seq model formulates its generative loss as 
% \begin{equation}
% % \tabularnewline
% \mathcal{L}_{\mbox{\tiny{seq2seq}}}(\theta) = -\log P(Y|X) = - \sum\nolimits_{t=1}^{L} \log P\left(y_{t} | y_{1: t-1}, X\right).
% \end{equation}
% where  $Y=\{y_i\}_{i=1}^L$. For CTC, it  locates on  the top of the encoder such that the hidden state $h_i$ of encoder is projected to the CTC output $o_i$.
CTC is on the top of the encoder and is jointly trained with the Seq2Seq model. 
% The loss of CTC is defined as:
% \begin{equation}
% % \tabularnewline
% \mathcal{L}_{\mbox{\tiny{ctc}}}(\theta) = - \log P(Y|X) = - \log \sum\nolimits_{\pi \in \Omega(Y)}{P(\pi|X)} \approx- \log \sum\nolimits_{\pi \in \Omega(Y)}{\prod\nolimits_{i=1}^{L}{P(o_i|X)} },
% \end{equation}
% where the set $\Omega(Y)$ denotes all possible sequences $\pi$ obtained by arbitrarily repeating symbols  $Y$ and inserting blank symbols into $Y$. Then ASR network combines these two components and minimizes 
Then ASR network combines these two components and minimizes $\mathcal{L}(\theta) = \lambda_{\mbox{\tiny{ctc}}} \mathcal{L}_{\mbox{\tiny{ctc}}}+\left(1-\lambda_{\mbox{\tiny{ctc}}}\right) \mathcal{L}_{\mbox{\tiny{seq2seq}}}$. 
% More details please refer to Sec.~\ref{low_asr_setup}.
% \begin{equation}
% % \tabularnewline
% \mathcal{L}_{\mbox{\tiny{asr}}}(\theta) = \lambda_{\mbox{\tiny{ctc}}} \mathcal{L}_{\mbox{\tiny{ctc}}}+\left(1-\lambda_{\mbox{\tiny{ctc}}}\right) \mathcal{L}_{\mbox{\tiny{seq2seq}}}.
% \label{eq:asr_loss}
% \end{equation}

%\subsubsection{Multilingual ASR model}
 \noindent\textbf{Multilingual ASR model.} 
To overcome the challenges brought by different sub-word units, lexicon and word inventories between different languages, we take the union over all the language-specific token sets and train a single model on a mixture dataset which combines all the source language data~\cite{Watanabe2017LanguageIE,Toshniwal2018MultilingualSR}. 
% To efficiently utilize different sub-word units, lexicon and word inventories between different languages, we take the union over all the language-specific token sets and train a single model on combined dataset from all the source languages like~\cite{Watanabe2017LanguageIE,Toshniwal2018MultilingualSR}. 
% Given $N$ languages with training sets $\{D_1, D_2, ... , D_N\}$ and token sets $\{C_1, C_2, ... , C_N\}$, 
Given $N$ languages with training sets $\{D_i\}_{i=1}^N$ and token sets $\{C_i\}_{i=1}^N$, the mixture training set is $\mathcal{D}_{\mbox{\tiny{multilingual}}} = \cup_{i=1}^N D_i$ and the token set for the mixture dataset is $\mathcal{C}_{\mbox{\tiny{multilingual}}} = \cup_{i=1}^N C_i$.

%In this work, we and allows all of the model parameters to contribute to handling the variations between different languages.
% Compared with ``multilingual bottleneck'' models with shared hidden layers and many language-specific heads~\cite{Dalmia2018SequenceBasedML,Grzl2014AdaptationOM,Fr2015MultilingualBF}, our architecture allows all of the model parameters to contribute to handling the variations between different languages and thus is simpler .

\subsection{Multilingual Meta-learning ASR}\label{ASRML}
% \noindent\textbf{Multilingual Meta-learning ASR model.}
% 为啥用meta learning，有啥好处，为什么可以work
% 直接用multitask learning pretraining多个source tasks，容易让初始化参数学习到source tasks上，且倾向于large task，不适用于adaptation。
% meta learning在meta training的时候就将adaptation考虑进来，目的是adaptation后的结果好。
% 因此meta learning很适用于我们的low resource asr的setting，能够通过很多其他语种学习到较好的初始化参数，再只需要low resource target语种的少量数据就可以很快速地adapt。
%Multilingual ASR learns parameter  initialization  via multi-task pre-training on several source tasks to alleviate the data  requirement of target task.  But the learnt  initialization  tends to overfit the source tasks with large amounts of data and underfit the source tasks with few data, leading a biased initialization and  hindering the adaptation for target language.

%Meta-learning has shown its superiority in few-shot learning  tasks~\cite{Finn2017ModelAgnosticMF,Nichol2018OnFM}, which learns a good parameter initialization such that the initialization can fast adapt to all source tasks via a few gradient steps on a few data. %So the learnt parameter initialization are often close to all source tasks. %Since the source and target tasks comes from the same task distribution,  the parameter initialization also only requires a few data to adapt  to target  task. 
%So we exploit meta learning to build our low-resource ASR model similar to~\cite{Hsu2019MetaLF}.
We train a multilingual meta-learning ASR (\textbf{MML-ASR}) model   on all languages to pursue the few-shot learning ability to  handle the low resource recognition problems.  Specially,  we use $f(\theta)$  to denote a multilingual ASR model parameterized by  $\theta$  and adopt  $\mathcal{D}_{\mbox{\tiny{source}}} = \{D_{\mbox{\tiny{source}}}^k\}_{k=1}^K$ to denote $K$ kinds of  source languages. To apply meta learning, \eg~ MAML~\cite{Finn2017ModelAgnosticMF} and Reptile~\cite{Nichol2018OnFM},  for the $k$-th kind of  language, we   sample partial data $D_{\mbox{\tiny{task}}}^k$  (a few sentences)  from $D_{\mbox{\tiny{source}}}^k$ to construct a small  recognition \textbf{task} $\mathcal{T}_k^i$. Then we  split   $D_{\mbox{\tiny{task}}}^k$ into \textbf{support data} $D_{\mbox{\tiny{support}}}^{k}$ and \textbf{query data} $D_{\mbox{\tiny{query}}}^{k}$. Accordingly, for each language we can sample many tasks denoted by $\mathcal{T}_{k}=\{\mathcal{T}_{k}^{i}\}_{i=1}^{n_k} $, where $n_k$ is the \textbf{task quantity} of the $k$-th kind of  language. Let $V$ be the total number of examples in the $k$-th kind of  language and $w$ be the number of examples per task, then $n_k$ could be calculated by combination number $C_V^w$.
For brevity, let $\mathcal{T}$ be the all task set $\mathcal{T}=\{\{\mathcal{T}_{1}^{i}\}_{i=1}^{n_1}, \cdots, \{\mathcal{T}_{K}^{i}\}_{i=1}^{n_K}\} $.  
% Now we can formulate our multilingual meta learning  ASR:
Now MML-ASR can be formulated as:
\begin{equation}\label{eq:meta-objective}
\min\nolimits_{\theta}\ \mathbb{E}_{\mathcal{T}_i\sim \mathcal{T}} \mathcal{L}_{D_{\mbox{\tiny{query}}}} (\theta - \alpha \nabla_\theta \mathcal{L}_{D_{\mbox{\tiny{support}}}}(\theta)),
\end{equation}
%\begin{equation}
%\mathcal{L}_{\mathcal{D}_{\mbox{\tiny{source}}}}^{\mbox{\tiny{meta}}}(\theta) = \mathbb{E}_{k\sim \mathcal{D}_{\mbox{\small{\mbox{\tiny{source}}}}}}\mathbb{E}_{D_{\mbox{\tiny{support}}}^{k}, D_{\mbox{\tiny{query}}}^{k}}[\mathcal{L}_{D_{\mbox{\tiny{query}}}^{k}}(\theta_k^{(t)})].
%\end{equation}
where $\mathcal{T}_i$ is sampled from $ \mathcal{T}$, $D_{\mbox{\tiny{support}}}$, and $D_{\mbox{\tiny{query}}}$ respectively denote the support and query data in task $\mathcal{T}_i$.  This model can be understood as that given a common model parameter $\theta$ for all language (or tasks),  for a sampled task $\mathcal{T}_i$, we  adapt the  parameter $\theta$ to this specific task via running one gradient descent on its support data  and obtain the task specific model parameter $\theta_{\mathcal{T}_i}=\theta - \alpha \nabla_\theta \mathcal{L}_{D_{\mbox{\tiny{support}}}}(\theta)$. Then we  evaluate the effectiveness of $\theta_{\mathcal{T}_i}$ on the query data $D_{\mbox{\tiny{query}}}$ of task   $\mathcal{T}_i$ and use this \textbf{query loss}, denoted as $Q_{\mathcal{T}_i}=\mathcal{L}_{D_{\mbox{\tiny{query}}}}(\theta_{\mathcal{T}_i})$, to guide the learning of the model parameter $\theta$.  %See optimization details in   MAML~\cite{Finn2017ModelAgnosticMF}, First-Order MAML~\cite{Finn2017ModelAgnosticMF} and Reptile~\cite{Nichol2018OnFM}. 
This process actually requires the learnt common model parameter $\theta$ to be close to the optimal model of all tasks $\mathcal{T}_i$ in task set $\mathcal{T}$ such that taking only one gradient step on a small-sized dataset  $D_{\mbox{\tiny{support}}}$ can achieve satisfactory performance on the query data $D_{\mbox{\tiny{query}}}$.  This mechanism gives the few shot learning  ability of model parameter $\theta$. 

After training, given a new speech recognition task with  a few training data, we  adapt  the model parameter $\theta$ to this  task by  a few gradient descent steps and obtain a task specific model  for test. So  this  MML-ASR model can well handle the low resource speech recognition  problem.% thanks to its few shot learning ability. 

This MML-ASR is inspired by  the prior meta-learning  ASR framework~\cite{Hsu2019MetaLF}.  Hsu~\et~\cite{Hsu2019MetaLF} first used a shared backbone to extract common features for all  languages  and then adopted different network branches to learn language-specific features, while MML-ASR here uses an entire shared model for all languages to simplify operations and make full use of information from different languages.
% But MML-ASR differs the specific  branch still requires sufficient language data for training, which is indeed restrictive for minority and isolated languages. By comparison, our totally shared model is  simpler and can get rid of this issue, since it enable the model to utilize the information from different languages better and have stronger ability to handle the variations among them. 

% \subsection{Adversarial Policy Network}
% \subsection{Multilingual Adversarial Meta Learning}
% \vspace{-0.4em}
\section{Adversarial Meta Sampling}
\begin{figure*}
	\centering
	\includegraphics[width=0.9\linewidth]{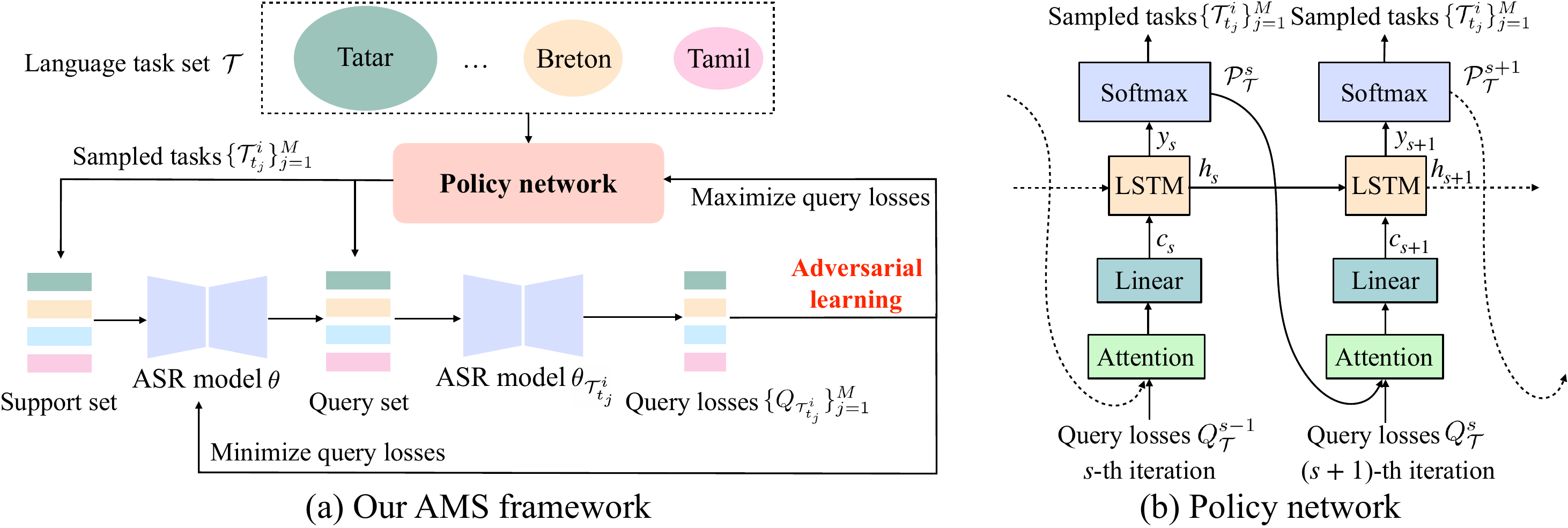}
% 	\vspace{-2mm}
	\caption{(a) The illustration of our AMS framework. (b) The architecture of the policy network.}
	\label{fig:overview}
% 	\vspace{-2.5mm}
    % \vspace{-2mm}
\end{figure*}

\subsection{Motivation}
\label{sec:aversarial}
% 存在task imbalance问题
As introduced in Sec.~\ref{ASRML}, for each meta-training iteration, we need to sample a task  $\mathcal{T}_i$ from the task set $\mathcal{T}=\{\{\mathcal{T}_{1}^{i}\}_{i=1}^{n_1}, \cdots, \{\mathcal{T}_{K}^{i}\}_{i=1}^{n_K}\} $ where $\mathcal{T}_{k}=\{\mathcal{T}_{k}^{i}\}_{i=1}^{n_k}$ denotes the tasks in the $k$-th language domain.  In real-world scenarios, different languages have a diverse geographic location, phonology, phonetic inventory, language family, and orthography, and their datasets vary greatly in size. So when sampling task $\mathcal{T}_i$  from all languages task set $\mathcal{T}$  to train our MML-ASR model, there are two severe issues. The first one is that the underlying  task quantity ($n_1,\cdots,n_K$) of each language  fluctuates  over a wide range, which means that the task sets  $\mathcal{T}_{k}\ (k=1,\cdots, K)$ are imbalanced in terms of their task quantity (\textbf{task-quantity imbalance}).  The second issue is that  the  tasks sampled from different  language task sets $\mathcal{T}_{k}\ (k=1,\cdots, K)$  actually have very different recognition difficulty due to the aforementioned language specificities~\cite{Waibel2000}, leading to an imbalance in terms of task difficulty (\textbf{task-difficulty imbalance}). 
% \pz{It is better to give one example to show different  task difficulty.} 
So sampling approaches become especially important in  ASR. Note, uniform sampling  or task-quantity-balanced sampling, namely sampling rate for each language positively relying on its task quantity in $\mathcal{T}_{k}$, usually ignore the task-difficulty  imbalance, and cannot achieve satisfactory performance as shown in our experiments.

%In real-world scenarios, different languages have diverse geographic location, phonology, phonetic inventory, language family and orthography, which leads to recognition challenges and the training samples of each language may also vary widely. However, ignoring such task imbalances, the traditional meta-learning methods random sample the source tasks for each meta-training step under a uniform distribution while each sampled task has equal training examples. In consequence, the sampling rate of the large task is so smaller than the one of the small task that the large task is easy to forget by the model and the initialization parameters may bias to the small task. 
%Intuitively, we can sample tasks according to their sizes to ensure that the sampling rates of all tasks are the same. However, beside the quantity imbalance, the other imbalances like language diversity cannot be eliminated by manual strategies. Therefore, an automatic task sampling method is desired to alleviate the task-imbalance problem in meta-learning.

% 为了解决提出了sampling机制
% 1. 发现query loss measure imbalance.
To resolve above two imbalance issues, we propose a novel and effective adversarial meta sampling approach that adaptively determines the sampling probability for each language task set  $\mathcal{T}_{k}$ in the meta-training process to balance both task quantity and difficulty in different language domains. Specifically, we observe that the query losses of tasks can well measure both imbalances because that if one language task set $\mathcal{T}_{k}$ are not well sampled in terms of its task quantity and task difficulty, then  its tasks have relatively large query loss. 
%\pz{I just revise here. In the following context, you should introduce your policy network, why use it? what architecture? what input and output? how does it integrate with Meta learning?} 
% 2. 直观地想，采用query loss直接在sample，但有坏处：a b
% 3. 因此我们设计一个基于LSTM和attention的网络，自动记录历史loss，和自动学习历史query loss当前query loss，与sample task的关系，解决了坏处 a b
% 4. 除此之外，我们引入了adversarial learning来达到sample query loss大的需要更多训练的task，他是最大化*** 最小化
% 5. 通过这样online和对抗的方式，sample policy可以在训练中动态改变，从而让更多低采样率或高难度的task得到充分的训练，也让asr模型变得更加有效。
% 6. 在ablation实验中，我们比较了上面说过的所有采样方式，证明了我们的方法效果最好。同时我们通过分析sampled task，发现我们方法确实sample出更多low sample rate和difficult task，解决了task imbalance问题，as shown in ***.
%\xyb{introduce policy network and why use}
% Intuitively, we can directly sample tasks based on the proportion of query losses of the current step, but it has two problems. 1) The unsampled tasks of the current step has no query loss. 2) It ignores the long-term information which may lead to local optimum. 
Intuitively, at each iteration, one can sample several tasks from each language tasks $\mathcal{T}_{k}$ and compute the average query loss $Q_{\mathcal{T}_{k}}$ for each $\mathcal{T}_{k}$. Then a simple way is to use $Q_{\mathcal{T}_{k}}/\sum_{i=1}^{K} Q_{\mathcal{T}_{k}}$ as the sampling probability for each $\mathcal{T}_{k}$. However, this method  ignores the long-term query loss information and only greedily assigns a large probability to some language tasks $\mathcal{T}_{k}$ according to the current query loss, which could be too locally greedy and leads to performance degradation.   Maintaining  a query loss buffer which linearly or exponentially averages the historical query losses still cannot achieve  satisfactory performance,  since the importance of current query loss and historical query losses is not necessarily a simple (exponential) average relation. Moreover, it requires to tune several manual hyper-parameters (\eg, window size) for computing average query losses.  All sampling methods mentioned above have been evaluated in Table~\ref{tab:Diversity9}.

% \vspace{-0.4em}
% \vspace{-2mm}
\subsection{Our Sampling Approach}
\label{sec:sampling_approach}

\begin{algorithm*}[h!]
    \small
  \caption{Adversarial Meta Sampling}
  \begin{algorithmic}[1]
    \Require $\alpha, \beta, \gamma$: step size hyperparameters
    \State Initialize $\theta$, $\phi$
    \State Initialize  $Q_{\mathcal{T}_{k}} = 0, \forall k \in \{1,2,...,K\}$
    \While {not done}
    \State Generate $K$-dim vector of sampling probabilities $\mathcal{P}_{\mathcal{T}} = (\mathcal{P}_{\mathcal{T}_1}, \mathcal{P}_{\mathcal{T}_2}, ... , \mathcal{P}_{\mathcal{T}_K})$ using $f_\phi$
    \State Sample $M$ language task set $\{\mathcal{T}_{t_j}\}_{j=1}^{M}, t_j\!\in\!\{1,\cdots\!, K\}$ with top-$M$ largest sampling probabilities 
    \State Sample one task $\mathcal{T}_{t_j}^i$ from each $\mathcal{T}_{t_j}$ to form $\{\mathcal{T}_{t_j}^i\}_{j=1}^{M}, t_j \in \{1,\cdots, K\}$ for meta-training

      \For {all ${\mathcal{T}}_{t_j}^i$} 
      \State Generate support set $D_{\mbox{\tiny{support}}}^{t_j}$ and query set $D_{\mbox{\tiny{query}}}^{t_j}$  from ${\mathcal{T}}_{t_j}^i$
      \State Compute adapted parameters with respect to $D_{\mbox{\tiny{support}}}^{t_j}$ using $\theta_{{\mathcal{T}}_{t_j}^i}=\theta - \alpha \nabla_\theta \mathcal{L}_{D_{\mbox{\tiny{support}}}^{t_j}}(\theta)$
      %\State Generate query set $D_{\mbox{\tiny{query}}}^{t_j}$ from ${\mathcal{T}}_{t_j}^i$ for the meta-update
      \EndFor
    %   \State Update $\theta$ with respect to $D_{\mbox{\tiny{query}}}^{t_j}, \forall j \in \{1,2,...,M\}$ using $Q_{{\mathcal{T}}_{t_j}^i}$
      \State Update $\theta \gets \theta - \beta \nabla_\theta \sum_{j=1}^M Q_{\mathcal{T}_{t_j}^i}$, where $Q_{\mathcal{T}_{t_j}^i} = \mathcal{L}_{D_{\mbox{\tiny{query}}}^{t_j}}(\theta_{{\mathcal{T}}_{t_j}^i})$ using each $D_{\mbox{\tiny{query}}}^{t_j}$
      \State Update query loss $Q_{\mathcal{T}} \!=\! (Q_{\mathcal{T}_{1}},   ... ,Q_{\mathcal{T}_{K}})$, where $Q_{\mathcal{T}_{k}} \!=\! Q_{\mathcal{T}_{t_j}^i|_{t_j=k}}$ if $\mathcal{T}_{k}$ is sampled else $Q_{\mathcal{T}_{k}} \!=\! Q_{\mathcal{T}_{k}}^{s-1}$
      \State Update $\phi \gets \phi + \gamma {\nabla}_{\!\phi}  \mathsmaller{\sum\nolimits_{j=1}^{M}}\! \mathcal{P}_{\mathcal{T}_{t_j}} \mathcal{L}_{D_{\mbox{\tiny{query}}}^{t_j}} (\theta - \alpha \nabla_\theta \mathcal{L}_{D_{\mbox{\tiny{support}}}^{t_j}}(\theta))$
    \EndWhile
  \end{algorithmic} 
  \label{alg1}
\end{algorithm*} 

To solve the above issues, 
% we propose a novel Adversarial Meta Sampling framework to improve multilingual meta-learning ASR models by adaptively sampling tasks in meta-training process to balance the diversity of different language tasks. 
we propose adversarial meta sampling method by designing a policy network which injects  attention mechanism  into LSTM~\cite{Hochreiter1997LongSM}. At  each training iteration,  it can adaptively predict  the most befitting probability to sample from each language tasks  $\mathcal{T}_{k}$ by using the long-term information in LSTM and the current query losses. Moreover, the policy network can be jointly trained with MML-ASR model in an end-to-end way without manual tuning extra hyper-parameters.

Specifically, as shown in Fig.~\ref{fig:overview} (a), at each meta-training iteration, the policy network samples $M$ kinds of language task set denoted by $\{\mathcal{T}_{t_j}\}_{j=1}^{M}, t_j \in \{1,\cdots, K\}$ from the $K$ kinds of language task $\mathcal{T}$ and then samples one task $\mathcal{T}_{t_j}^i$ from each $\mathcal{T}_{t_j}$ to form training task set $\{\mathcal{T}_{t_j}^i\}_{j=1}^{M}$ for meta-training of MML-ASR model. So the meta-objective of MML-ASR model in Eqn.~\eqref{eq:meta-objective} can be reformulated as
\begin{equation}
\min\nolimits_{\theta}\ \mathbb{E}_{\pi \sim f_\phi} \mathbb{E}_{\mathcal{T}_i\sim \pi(\mathcal{T})} \mathcal{L}_{D_{\mbox{\tiny{query}}}} (\theta - \alpha \nabla_\theta \mathcal{L}_{D_{\mbox{\tiny{support}}}}(\theta)),
\label{eq:meta_obj}
\end{equation}
where  $\pi$ denotes the task sampling policy   learnt by  the  policy network $f_\phi$ parameterized by  $\phi$.

After meta-training of MML-ASR model, we can obtain the query losses $\{Q_{\mathsmaller{\mathsmaller{\mathcal{T}_{t_j}^i}}}\}_{j=1}^M$ of each training task $\{\mathcal{T}_{t_j}^i\}_{j=1}^{M}$. As mentioned in Sec.~\ref{sec:aversarial},  the query losses of tasks drawn  from each language task set $\mathcal{T}_{k}$  can well measure the task imbalances in terms of both task quantity and task difficulty.  This is because if one language task set $\mathcal{T}_{k}$ are not well sampled  in terms of its task quantity and task difficulty, then  its tasks have relatively large query loss.  This actually means that  the language  task set  $\mathcal{T}_{k}$ with large query loss $Q_{\mathcal{T}_{k}}$  requires  more extra training.  So at each iteration,  our policy network  attempts to increase the query loss of MML-ASR model through adversarial learning for sampling the proper language  task set for training. Formally, the objective loss of our policy network   is defined  as:
\begin{equation}
    \begin{split}
        &\phi^{\ast} = \mathop{\arg\max}\nolimits_{\phi} \mathcal{J}(\phi), \text{where} \mathcal{J}(\phi) = \\
        \mathbb{E}_{\pi \sim f_\phi}& \mathbb{E}_{\mathcal{T}_i\sim\pi(\mathcal{T})} \mathcal{L}_{D_{\mbox{\tiny{query}}}} (\theta - \alpha \nabla_{\theta} \mathcal{L}_{D_{\mbox{\tiny{support}}}}(\theta)).
    \end{split}
\label{eq:policy_obj}
\end{equation}
%$f(\phi)$  denotes the policy network parameterized by  $\phi$ and $\pi(\cdot)$  the task distribution over source tasks $\mathcal{T}$ sampled by policy network $f_\phi$.

Then we focus on introducing the  policy network $f_\phi$, which shown in Fig.~\ref{fig:overview} (b).  First, we use  $K$-dim vector $Q_{\mathcal{T}}^{s-1} = (Q_{\mathcal{T}_{1}}^{s-1}, Q_{\mathcal{T}_{2}}^{s-1}, ... ,Q_{\mathcal{T}_{K}}^{s-1})$ to denote the query loss for each language task set $\mathcal{T}_{k}$ at the $(s-1)$-th meta-training iteration. Second, at the $s$-th  iteration, the policy network will  output a $K$-dim   $\mathcal{P}_{\mathcal{T}}^{s} = (\mathcal{P}_{\mathcal{T}_1}^{s}, \mathcal{P}_{\mathcal{T}_2}^{s}, ... , \mathcal{P}_{\mathcal{T}_K}^{s})$ in which   $\mathcal{P}_{\mathcal{T}_k}^{s}$ denotes the sampling probability for the $k$-th language task set $\mathcal{T}_k$. Third, as aforementioned, we select the top-$M$ sampling probabilities and respectively sample one task from their  corresponding language task set $\mathcal{T}_k$ for meta-training. Then we use the $M$ new query loss $\{Q_{\mathsmaller{\mathsmaller{\mathcal{T}_{t_j}^i}}}^{s}\}_{j=1}^M$ to update the corresponding query loss in  $Q_{\mathcal{T}}^{s-1} $ for obtaining $Q_{\mathcal{T}}^{s} $.

Next, at the $(s+1)$-th iteration, we feed $Q_{\mathcal{T}}^{s} $ and $\mathcal{P}_{\mathcal{T}}^{s} $  into the policy network and combine these two inputs ($Q_{\mathcal{T}}^{s}, \mathcal{P}_{\mathcal{T}}^{s}$) to calculate the feed forward attention, and then get the attention output $c_{s+1}$ through a fully-connected layer. Then, a LSTM layer takes the hidden state of previous LSTM cell $h_{s}$ as well as attention output $c_{s+1}$ as input, and generates the LSTM output $y_{s+1}$ and current hidden state $h_{s+1}$.  Finally, based on $y_{s+1}$, we use fully-connected layer with Softmax function to predict  a probability vector $\mathcal{P}_{\mathcal{T}}^{s+1} = (\mathcal{P}_{\mathcal{T}_1}^{s+1}, \mathcal{P}_{\mathcal{T}_2}^{s+1}, ... , \mathcal{P}_{\mathcal{T}_K}^{s+1})$.  In this way, same as the $s$-th iteration,  we can select the top-$M$ largest probabilities and   sample tasks from their corresponding task sets for meta-training.

%
%
%The first one is that the underlying  task number ($n_1,\cdots,n_K$) of each language  fluctuates  over a wide range, which means that the task sets  $\mathcal{T}_{k}\ (k=1,\cdots, K)$ are imbalanced in terms of their task quantity.  The second issue is that  the  tasks sampled from different  language task sets $\mathcal{T}_{k}\ (k=1,\cdots, K)$  actually have very different recognition difficulty due to the aforementioned language specificities~\cite{Waibel2000}, leading to an imbalance in terms of task difficulty.  
%Besides, the $K$-dim query losses vector $Q_{\mathcal{T}} = (Q_{\mathcal{T}_{1}}, Q_{\mathcal{T}_{2}}, ... ,Q_{\mathcal{T}_{K}})$ and the probability vector $\mathcal{P}_{\mathcal{T}} = (\mathcal{P}_{\mathcal{T}_{1}}, \mathcal{P}_{\mathcal{T}_{2}}, ... ,\mathcal{P}_{\mathcal{T}_{K}})$ of the corresponding tasks $\{\mathcal{T}_{k}\}_{k=1}^K$ are also used to update the policy network. 

As the discrete sampling operations for obtaining  $M$ tasks is not differentiable, we apply REINFORCE algorithm~\cite{Williams1992SimpleSG} to solve this issue and optimize the policy network via the following gradient, 
\begin{equation*}
\begin{aligned}
{\nabla}_{\!\phi} \mathcal{J}(\phi) &\!=\! {\nabla}_{\!\phi} \mathbb{E}_{\pi \sim f_\phi} \mathbb{E}_{\mathcal{T}_i\sim \pi(\mathcal{T})} \mathcal{L}_{D_{\mbox{\tiny{query}}}} (\theta - \alpha \nabla_\theta \mathcal{L}_{D_{\mbox{\tiny{support}}}}(\theta)) \\
% &\approx \sum_m{\mathcal{L}_m}{\nabla}_{\phi} p_m = \sum_m{\mathcal{L}_m}p_m{\nabla}_{\phi}\log{p_m} \\
&\!\approx\! 
%\mathop{\mathbb{E}}\nolimits_{\pi \sim f_\phi}
 {\nabla}_{\!\phi}  \mathsmaller{\sum\nolimits_{i=1}^{M}}\! \mathcal{P}_{\mathcal{T}_i} \mathcal{L}_{D_{\mbox{\tiny{query}}}} (\theta - \alpha \nabla_\theta \mathcal{L}_{D_{\mbox{\tiny{support}}}}(\theta)).
%{Q_{\mathcal{T}_{i}}}{\nabla}_{\phi}
%\log{\mathcal{P}_{\mathcal{T}_i}}  % \approx {{\nabla}_{\phi}{Q_{\mathcal{T}}}\log{\mathcal{P}_{\mathcal{T}}}} .
\end{aligned}
\end{equation*}
%Through the derivation of Section~\ref{sec:joint_train}, the training procedure of the policy network can be formulated as,
%\begin{equation}
%\begin{split}
%{\nabla}_{\phi} \mathcal{J}(\phi)  \approx {{\nabla}_{\phi}{Q_{\mathcal{T}}}\log{\mathcal{P}_{\mathcal{T}}}}, \quad \phi  \gets \phi + \gamma {{\nabla}_{\phi}{Q_{\mathcal{T}}}\log{\mathcal{P}_{\mathcal{T}}}}.
%\end{split}
%\label{eq:update_policy}
%\end{equation}
Through such an online and adversarial manner, the sampling policy  is dynamically changed along with the training state of the MML-ASR. In this way,  the language task set $\mathcal{T}_{k}$ that is not well sampled  in terms of its task quantity and task difficulty will be sampled more for more effective  learning. 

Moreover, \textit{our sampling method can be applied to not only MML-ASR methods but also multilingual transfer learning ASR (\textbf{MTL-ASR}) without any architecture modification}. For MTL-ASR, we  directly use the training loss $\mathcal{L}_{\mathcal{T}_{k}}$ of a task sampled from each source language domain to construct the loss vector $ (\mathcal{L}_{\mathcal{T}_{1}}, \mathcal{L}_{\mathcal{T}_{2}}, ... ,\mathcal{L}_{\mathcal{T}_{K}})$ and feed it into  our policy network. Similarly, the policy network outputs the sampling probability for each language domain. Experimental results in Sec.~\ref{sec:mtl_result} also verify the superiority of our policy network and show significant improvement when applying our sampling method into MTL-ASR.
Finally, our Adversarial Meta Sampling framework is summarised in Algorithm ~\ref{alg1}.

\section{Experiments}\label{experiments}
% \subsection{Multilingual Low-resource ASR}
% \subsection{Experimental Setup}
\noindent\textbf{Datasets.}
% 首先，为了模拟task imbalance问题，我们从common voice里面随机选择了11种种类和数据量都差异很大的语言，9种作为source task，2种作为target task，组成Diversity11数据集，详见table。
% 其次，为了做对比，我们又构造了一个差异较小的多语言数据集Indo12，它里面也包含9个source task，都来自印欧语系且数据均为10小时除了swedish，然后也选取了两个印欧语系的语种作为target task，除此之外，还选择了一个非印欧语系的target语种作为对比，详见table。
% 最后为了证明我们方法能够在更少的source task上有效，我们从Indo12中去除了3中source task，russian, swedish, welsh，组成了Indo6数据集。
Common Voice~\cite{commonV} is an open-source multilingual voice dataset and contains about 40  kinds of  languages. For low-resource evaluation,  we construct  three different datasets: Diversity11, Indo12, and Indo9 which are described in  Table~\ref{tab:dataset}. To construct the Diversity11 dataset, we randomly select 11 kinds of languages from different districts with varying diversities and quantities, and divide them into 9 source languages and 2 target languages. For Indo12, we  randomly select  11 kinds of languages with little difference from the same Indo-European language family and 1 Afro-Asiatic language. To test our method on fewer source languages, we remove 3 languages (Russian, Swedish, and Welsh) from Indo12 to obtain Indo9. 

In addition, we also conducted experiments on the IARPA BABEL dataset~\cite{Gales2014SpeechRA} with 6 source languages (Bengali, Tagalog, Zulu, Turkish, Lithuanian, Guarani) and 3 target languages (Vietnamese, Swahili, Tamil).

% \vspace{-1em}
\begin{table} 
% \vspace{-2.2em}
\scriptsize
\centering
\setlength{\tabcolsep}{3.0pt}  

% \vspace{-0.4em}
\begin{tabular}{c|c|cc|cc|cc}
\toprule
\multirow{5}{*}{Diversity11} & \multirow{3}{*}{Source} & Turkish            & 13            & Tatar     & 25            & Tamil             & 3   \\ \cline{3-8} 
                             &                               & Swedish            & 5             & Mongolian & 9             & Latvian           & 4   \\ \cline{3-8} 
                             &                               & Dhivehi            & 6             & Breton    & 5             & Arabic            & 7   \\ \cline{2-8} 
                             & \multirow{2}{*}{Target} & \multicolumn{2}{c}{Kyrgyz-\textit{train}}   & 10        & \multicolumn{2}{|c}{Kyrgyz-\textit{test}}   & 1   \\ \cline{3-8} 
                             &                               & \multicolumn{2}{c}{Estonian-\textit{train}} & 9        & \multicolumn{2}{|c}{Estonian-\textit{test}} & 1   \\ \hline \hline
\multirow{6}{*}{Indo12}      & \multirow{3}{*}{Source} & English            & 10            & Portuges  & 10            & Russian           & 10  \\ \cline{3-8} 
                             &                               & French             & 10            & German    & 10            & Welsh             & 10  \\ \cline{3-8} 
                             &                               & Italian            & 10            & Catalan   & 10            & Swedish           & 5   \\ \cline{2-8} 
                             & \multirow{3}{*}{Target} & \multicolumn{2}{c}{Spanish-\textit{train}}  & 10        & \multicolumn{2}{|c}{Spanish-\textit{test}}  & 1.5 \\ \cline{3-8} 
                             &                               & \multicolumn{2}{c}{Dutch-\textit{train}}    & 10        & \multicolumn{2}{|c}{Dutch-\textit{test}}    & 1.5 \\ \cline{3-8} 
                             &                               & \multicolumn{2}{c}{Kabyle-\textit{train}}   & 10        & \multicolumn{2}{|c}{Kabyle-\textit{test}}   & 1.5 \\ \bottomrule
\end{tabular}
\caption{Multilingual dataset statistics in terms of hours (h).}
% \vspace{-0.8em}
\label{tab:dataset}
% \vspace{-1em}
\end{table}

\begin{table*}[h]
	\centering
	\footnotesize
	\setlength{\tabcolsep}{5.0pt}  
	
% 	\vspace{-0.6em}
	\begin{tabular}{c|cc|cc|cc|cc}
		% \hline
		\toprule
		Target      & Kyrgyz & Estonian & \multicolumn{2}{c|}{Spanish} & \multicolumn{2}{c|}{Dutch}  &  \multicolumn{2}{c}{Kabyle}  \\ \hline
		Source &  \multicolumn{2}{c|}{\textit{Diversity11}} & \textit{Indo9} & \textit{Indo12} & \textit{Indo9} & \textit{Indo12} & \textit{Indo9} & \textit{Indo12}  \\ \hline
		Monolingual training~\cite{Hori2017AdvancesIJ}  & 76.25 & 86.04 & \multicolumn{2}{c|}{80.30} & \multicolumn{2}{c|}{68.71} & \multicolumn{2}{c}{85.41} \\
		TL-ASR~\cite{Kunze2017TransferLF}       & 68.28          & 82.04 &  \multicolumn{2}{c|}{79.39}  &  \multicolumn{2}{c|}{56.58} &  \multicolumn{2}{c}{89.12}  \\ \hline
		MTL-ASR (multi-head)~\cite{Dalmia2018SequenceBasedML}  & 67.56          & 81.50  & 75.82 & 73.00     & 57.80  & 56.41    & 82.10  & 81.23          \\
		MTL-ASR~\cite{Watanabe2017LanguageIE} & 64.90          & 83.70	 & 73.85 & 71.40        & 62.55  & 58.46      & 84.25    & 81.88          \\
		our AMS (MTL-ASR) & \textbf{59.55} & \textbf{79.33} & \textbf{71.02}& \textbf{68.97}& \textbf{58.22}& \textbf{54.96}& \textbf{83.90}& \textbf{79.26} \\ \hline
		% Joint training + Reptile~\cite{Nichol2018OnFM}  & 73.59    & -  & 61.87  & - & 91.20  & -  \\

		% Joint training + MAML~\cite{Finn2017ModelAgnosticMF}   & 71.30 & 66.24   & 54.33 & 51.74  & 79.49 & 73.52  \\ \hline
		MML-ASR~\cite{Hsu2019MetaLF}  & 58.29          & 79.66 & 66.75 & 65.24      & 53.33 & 52.56  & 79.45 & 75.96    \\
		{our AMS (MML-ASR)} & \textbf{50.72} & \textbf{72.26}  & \textbf{65.21} & \textbf{64.40} & \textbf{51.18} & \textbf{49.13} & \textbf{78.21} & \textbf{73.69} \\ 
		
		% \textbf{our AMS-MAML}  &-  & \textbf{64.26} &-  & \textbf{50.33} &-  & \textbf{72.64} \\
		\bottomrule
	\end{tabular}
	\caption{Results of low resource ASR on Diversity11, Indo12 and Indo9 in terms of WER (\%).}
	\label{tab:Indo}
% 		\vspace{-0.4em}
\end{table*}

\noindent\textbf{Implementation Details.}
\label{low_asr_setup}
We use the joint attention-CTC ASR model~\cite{Kim2017JointCB,Hori2017AdvancesIJ}  as our ASR model. The encoder contains a 6-layered  VGG~\cite{Simonyan2015VeryDC} extractor and 5 BLSTM~\cite{Graves2013HybridSR,Graves2014TowardsES} layers, each with 320-dimensional units per direction. Location-aware attention~\cite{Chorowski2015AttentionBasedMF} with 300 dimensions is used in our attention layer and the decoder is a single LSTM~\cite{Hochreiter1997LongSM} layer with 320 dimensions. We set $\lambda_{\mbox{\tiny{ctc}}}$ to 0.5. During inference, the greedy-search decoding is used to get the best hypothesis. Following~\cite{Hori2017AdvancesIJ}, we use 80-dimensional log Mel-scale filterbank coefficients with pitch features as the input features. 
%Google's SentencePiece toolkit~\cite{sentencepiece} is employed to process audio transcripts.
Byte pair encoding (BPE) compression algorithm (Sennrich et al. 2016) is employed to process audio transcripts.
All transcripts in the different multilingual datasets are used to train sub-word models separately based on the BPE algorithm. The transcripts are pre-tokenized to sequences of sub-word units (tokens) one-hot vectors using the sub-word models. The policy network contains a feed forward attention and a one-layer LSTM with the hidden size 100 and the input size 32. We use Adam  with an initial learning rate $\gamma=0.035$ and an entropy penalty weight $10^{-5}$ to train the policy network.  
%To choose the optimal $M$, we evaluate the performance of our AMS verse different $M$, where $M \in \{ 2, 3, 4, 5, 7, 9\}$, and finally set $M$ to 3.
%In the experiments, 
We set $M$ as  3 after searching the range  $M \in \{ 2, 3, 4, 5, 7, 9\}$ and set $w$ as 48, of which 24 examples are divided into support set and 24 examples into query set.

\subsection{Results on Low-resource ASR}

\noindent\textbf{Results on Diversity11.}
% Diversity 11 结果
% 首先，在所有方法里，我们的方法最好。由于target set数据量极少，所以他们直接训练效果很差，必须通过low resource方式帮助让他们提升效果。
% 其次，在multilingual 方法joint training上加入meta learning后，效果会比joint training下降**，因为meta learning在训练过程中考虑的adaptation，有助于缓解multitask方法overfit到source tasks上的问题，学到适应于adaptation的初始化参数进一步提升了效果。
% 然而，diversity11数据集存在较大的task imbalance问题，传统meta learning方法使用的均匀采样会导致大数据集遗忘，以及无法考虑各个task之间的差异性而应该针对性采样。而我们的方法基于query loss自动地adversarial sampling，让每个不同的task拥有不同的采样率，可以见图，在meta learning的基础上又进一步提升了。
% 在表格中，reptile的效果不太好，我们分析认为，reptile在每个step中更新梯度量较大，因此在task imbalance的情况下，受影响也较大。而在使用了adversarial sampling后，效果明显提升。
% kyrgyz的结果比estonian好很多，因为estonian数据量比kyrgyz少更多。
Table~\ref{tab:Indo} reports the results on Diversity11 in terms of word error rate (WER). For all target languages, our AMS significantly outperforms all previous methods. First, the performance of monolingual is poor without the help of source languages. Second, meta-learning decrease WER over 6\% thanks to its few-shot learning ability by learning better initialization parameters that enjoy fast adaptation ability. Moreover, by learning to sampling tasks for meta-learning, our AMS further improves the results over 7\% on this dataset which has large task imbalance.

\noindent\textbf{Results on Indo12 and Indo9.}
% Indo12和Indo9结果
% 在Indo12中，我们方法的结果也最好。说明我们在数据量大致均等的情况下，也可以通过adversarial sampling挑选出更好较难的task提升效果，同时Indo9的结果说明我们也可以在更少的source tasks上取得效果。
% kabyle的结果比spanish和dutch效果差很多，因为kabyle是一个亚非语系语言，而我们source tasks选择的是印欧语系，说明source tasks更能够帮助与target task相似的语言，但是非相似的语言也有提升效果。
As shown in Table~\ref{tab:Indo},  our method consistently achieves the state-of-the-art performance  on Indo12 which eliminates task-quantity imbalance and Indo9 which has much fewer source languages for training. This is because   our AMS  uses adversarial sampling to select better tasks for effective learning and well overcomes the task-difficulty imbalance issue. The results of Kabyle are much worse than that of Spanish and Dutch because Kabyle is an Afro-Asiatic language and all source languages are Indo-European language, which indicates that source languages from the same language family are more helpful for target languages.
% \noindent\textbf{Analysis of task imbalance.}
% sample tasks图像分析
% 通过比较图像a和b可以发现，使用了adversarial sampling后，在task imbalance差别更明显的数据集diversity11上，不同task的sample次数差异很大，而在indo12上，差异比diversity11小很多。 同时我们的方法在diversity11上提升的效果比indo12上更为明显，说明我们的方法更适宜于解决task imbalance数据集，自动调整采样率，从而提升效果。
% 除此之外，通过分析diversity11上sample task的结果，我们发现sample次数与task的数据量成较大的正相关。说明我们的方法确实在拉升大数据集的采样率，解决大数据的遗忘，增加大数据集的覆盖率。
% \noindent\textbf{Analysis of quantity imbalance.}
% By comparing Figure~\ref{fig:task_imbalance} (a) and (b), we can observe that after using adversarial sampling, the sample times of different tasks differs greatly on the Diversity11 where the task imbalance problem is more serious, while on Indo12, the difference of task sample times is much smaller. At the same time, the performance of our method on Diversity11 is better than that of Indo12, indicating that our proposed method is more suitable for solving the situation with large task imbalance problem and automatically adjusting the sampling rate of different tasks to improve the performance. In addition, through analyzing the results of the sample times on Diversity11, we find that the sample times of the task have a large positive correlation with the example numbers of the task. It shows that our method is indeed increasing the sampling rate of large tasks, so as to alleviate the harm of forgetting large tasks and avoid initialization parameters toward small tasks. 

\noindent\textbf{Results on IARPA BABEL.}
In order to further verify the effectiveness of our AMS, we also conducted experiments to compare with previous works on the IARPA BABEL, which is another public multilingual dataset. Table~\ref{tab:babel} reports the results on BABEL.
%in terms of character error rate (CER).
In addition to comparing with the baselines above, we also selected the results of recent papers for comparison. As can be observed, our AMS achieves the best results for all target languages, which can improve the performance of both MML-ASR and MTL-ASR with the proposed adversarial meta sampling method. It further demonstrates that the improvement of our AMS can be  easily reproduced on different multilingual low-resource ASR datasets.

\begin{table}
	\centering
	\scriptsize
	\tabcolsep 0.02in 
	
% 	\vspace{-0.6em}
	\begin{tabular}{l|ccc}
		\toprule
        Method  & Vietnamese & Swahili & Tamil \\ \hline
        % \tabincell{c}{Monolingual (MultiCTC)\\~\cite{Hsu2019MetaLF}} & 71.80 & 47.50 & 69.90 \\
        Monolingual (Multi-CTC)(Hsu et al. 2020) & 71.80 & 47.50 & 69.90 \\
        Monolingual (BLSTMP)~\cite{Cho2018MultilingualSS} & 54.30 & 33.10 & 55.30 \\
        Monolingual (VGG-Small)~\cite{Chen2020DARTSASRDA} & 46.00 & 39.60 & 57.90 \\
        Monolingual (VGG-Large)~\cite{Chen2020DARTSASRDA} & 48.30 & 38.30 & 60.10 \\
        Monolingual (Joint attention-CTC)~\cite{Hori2017AdvancesIJ} & 48.68 & 38.62 & 54.45 \\ \hline 
        % TL-ASR & 50.74 & x & x & x \\
        MTL-ASR (Multi-CTC)(Hsu el al. 2020) & 59.70 & 48.80 & 65.60 \\
        MTL-ASR (Joint attention-CTC)(Watanabe et al. 2017) & 47.17 & 34.10 & 51.17 \\ 
        our AMS (MTL-ASR) & \textbf{45.51} & \textbf{33.15} & \textbf{49.57} \\
        \hline
        MML-ASR (Multi-CTC)(Hsu et al. 2020) & 50.10 & 42.90 & 58.90 \\
        MML-ASR (Joint attention-CTC)(Hsu et al. 2020) & 45.10 & 36.14 & 50.61 \\
        our AMS (MML-ASR) & \textbf{43.35} & \textbf{32.19} & \textbf{48.56} \\
        \bottomrule
	\end{tabular}
	\caption{Results of low resource ASR on IARPA BABEL in terms of Character Error Rate (CER\%).}
	\label{tab:babel}
% 	\vspace{-0.4em}
\end{table}
% \vspace{-0.4em}
\subsection{Ablation Studies}
Considering the superior performance of MML-ASR over  MTL-ASR, all the following ablation experiments focus on AMS based on MML-ASR.
\begin{table}
	\centering
	\scriptsize
	
% 	\vspace{-0.4em}
	\tabcolsep 0.02in 
	\begin{tabular}{@{}l|ll@{}}
		\toprule
		Method                   & Kyrgyz          & Estonian        \\ \hline
		MML-ASR (Reptile)~\cite{Nichol2018OnFM}                   & 66.51          & 83.17          \\
		MML-ASR (FOMAML)~\cite{Hsu2019MetaLF}                     & 59.23          & 78.64          \\
		MML-ASR (MAML) (Uniform)~\cite{Hsu2019MetaLF}                         & 58.29          & 79.66          \\ \hline
		PPQ-MAML~\cite{Dou2019InvestigatingMA}                                     & 58.95          & 77.26          \\
		PPQL-MAML~\cite{Sun2018MetaTransferLF}                                    & 54.87          & 74.97          \\
		PPEAQL-MAML & 55.14          & 75.41          \\ 
		PPAQL-MAML & 53.15          & 73.33          \\
		our AMS-MAML w/o attention & 54.16        & 74.29       \\ \hline
		{our AMS-Reptile} & {59.30} & {78.49} \\
		{our AMS-FOMAML}  & {53.04} & {74.77} \\
		{our AMS-MAML}    & \textbf{50.72} & \textbf{72.26} \\ \hline
		{our AMS-MAML (80\% target)} & 59.02 & 75.87 \\
		{our AMS-MAML (50\% target)} & 70.27 & 81.97  \\ 
		{our AMS-MAML (20\% target)}  & 87.11 & 91.72 \\ \bottomrule
	\end{tabular}
	\caption{Ablation study results on Diversity11 in terms of WER (\%).}
	\label{tab:Diversity9}
% 	\vspace{-0.4em}
\end{table}

\noindent\textbf{Different meta-learning methods.}
% \pz{What is this? Remove it ???}
 Table~\ref{tab:Diversity9} shows that  our AMS can improve  all  meta-learning methods, including MAML (Hsu et al. 2020; Finn et al. 2017), FOMAML (Hsu et al. 2020; Finn et al. 2017), Reptile (Nichol et al. 2018). Among them,    AMS-FOMAML achieves  similar performance as AMS-MAML but has higher training efficiency. So in the other speech tasks, we focus on AMS- FOMAML.

\noindent\textbf{Different scale of training data.}
% 为了验证我们方法在更少数据下的有效性，我们将target task数据量进一步减少到80%,50%和20%，大约是7h,5h,1h。
% 可以观察到，随着数据量进一步减少，识别效果也逐渐变差。
% 但仅仅使用80%原来的数据，我们方法都能够比大多数baseline使用100%数据的效果好。同时使用50%的数据我们的方法也依旧能够比少量low resource方法优，也会比mono-lingual优。
% 直到20%仅有1小时训练数据效果才降到80%-90%，由于数据量过少。
To test our method when data are only a few, we  reduce the training data of target languages to 80\%, 50\% and 20\%. From Table~\ref{tab:Diversity9} and~\ref{tab:Indo},  one can observe that with 80\% training data, our method still works better than most baselines with 100\% training data, which testifies our method can  effectively alleviate the need of heavy annotated training data. 

\noindent\textbf{Comparison among different sampling methods.}
% 为了验证我们每个模块的有效性，我们与不同sample方式进行了比较。
% Uniform 均匀采样
% PPS ***
% PPQL ***
\label{ablation:sampling}
We further compare the performances of different sampling methods. 1) Sampling tasks uniformly (Uniform). 2) Sampling tasks with the probability proportional to the task quantity of each language  task set (PPQ)~\cite{Dou2019InvestigatingMA}. 3) Sampling tasks with the probability proportional to the query loss of each language  task set (PPQL)~\cite{Sun2018MetaTransferLF}. 4) Sampling tasks with the probability proportional to the average query loss of each language  task set with a  window (PPAQL). 5) Sampling tasks with the probability proportional to the exponential average query loss of each language  task set (PPEAQL). 6) Our AMS without attention layer (AMS w/o attention). 
From Table~\ref{tab:Diversity9}, one can find that (1) PPQ is slightly better than Uniform by considering the task-quantity imbalance; (2)  the methods that use query loss to sample tasks have greatly improved the performance and PPAQL is the best; (3) our AMS significantly surpasses other methods, which indicates that our policy network can effectively exploit the long-term and instant information to sample the most benefiting tasks in the training process.
\subsection{Generalization   analyses}
\noindent\textbf{AMS on multilingual transfer learning.}
\label{sec:mtl_result}
Our  sampling method can be generalized to MTL-ASR as mentioned in Sec.~\ref{sec:sampling_approach}. As shown in  Table~\ref{tab:Indo}, our AMS (MTL-ASR) outperforms  MTL-ASR  on all datasets. It  shows that our AMS can effectively improve both meta-learning methods and multilingual transfer learning methods by simply incorporating a policy network for adversarial sampling.

% \begin{wrapfigure}{r}{0.6\linewidth}
\begin{figure}
% 	\vspace{-0.2em}
	\centering
	\includegraphics[width=0.6\linewidth]{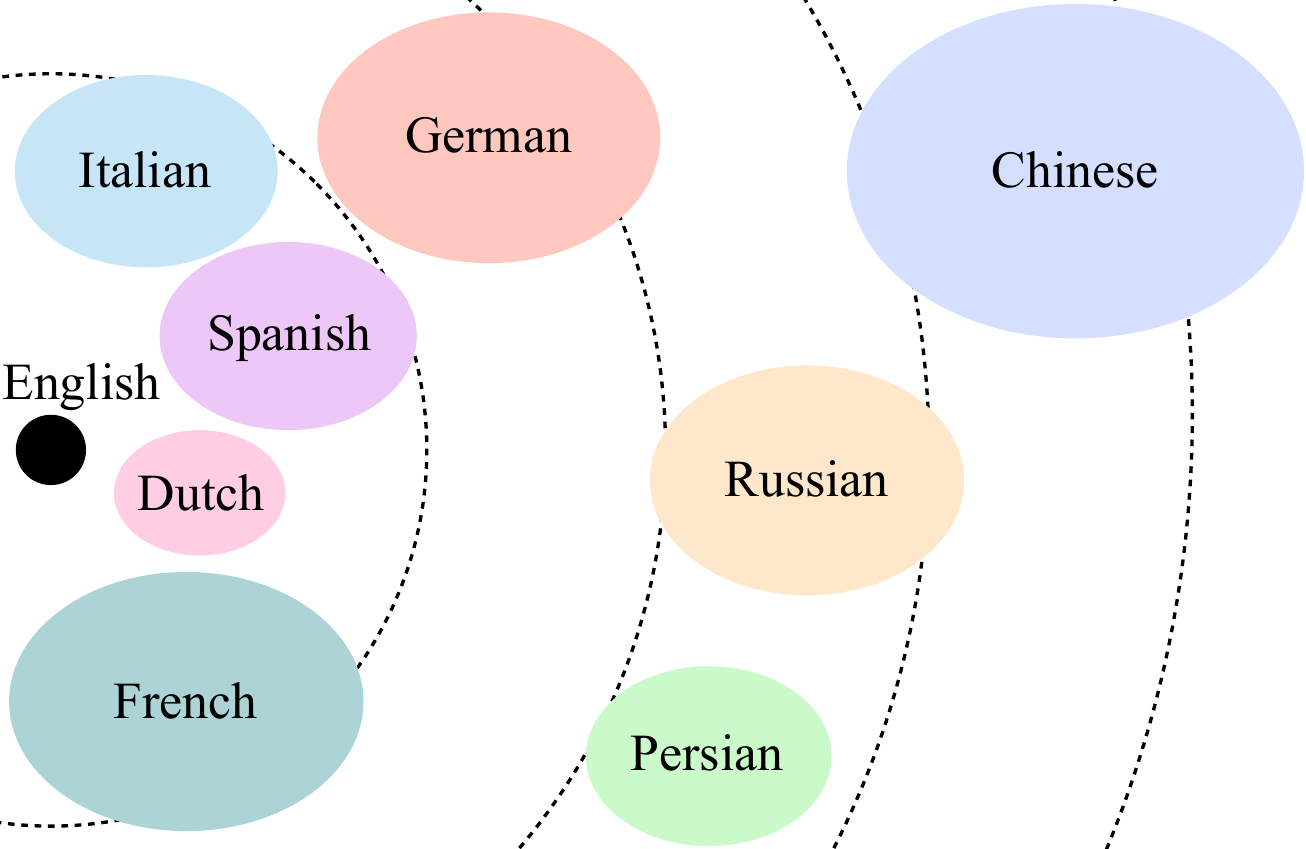}
	% \caption{Analysis of language difficulty (or relevance) from English to other languages. }
% 	\vspace{-0.8em}
	\caption{Imbalance analysis of task difficulty in our speech translation experiments. The distance from the dotted circle where the language is located to the black point represents the distance from the language to English and the circular area of the language represents the sample times of this language. }
	\label{fig:relation}
% 	\vspace{-0.4em}
\end{figure}

\noindent\textbf{AMS on speech classification.}
% \noindent\textbf{Datasets.} 
% Provided by the AutoSpeech 2020 competition~\cite{automlspeech}, our speech classification datasets have 5 public downloadable datasets and 5 non-downloadable feedback datasets, as source and target tasks respectively. Each provided dataset is from one speech classification domains, including speaker identification, emotion classification and so on. Moreover, in these datasets, the number of classes, the number of 
% samples, and the duration of samples vary greatly. As for the model, we use MobileNetV2 as feature extractor and NetVLAD as aggregation network to classify.
Our speech classification datasets contain 5 source datasets and 5 target datasets provided by the AutoSpeech 2020 competition\footnote{https://www.automl.ai/competitions/2}. Different datasets come from different speech classification domains with varying examples, classes and the quantity of examples, including speaker identification, emotion classification, etc. We evaluate our AMS with   MobileNetV2~\cite{Sandler2018MobileNetV2IR} as the feature extractor and NetVLAD~\cite{Arandjelovi2016NetVLADCA} as the aggregation network. As shown in Table~\ref{tab:speech_classification}, our AMS outperforms most of the baselines with large improvement.
%the number of classes varies from 2 to 500, while the number of instances varies from several to hundreds and the duration of instances is ranging from seconds to 1 minute.

\begin{table}[]
	\centering
	\scriptsize
	
% 	\vspace{-0.4em}
	\setlength{\tabcolsep}{2.6pt}  
	\begin{tabular}{l|ccccccc}
		\toprule
		Method  & \textit{D1}  & \textit{D2}  & \textit{D3} & \textit{D4}  & \textit{D5} & Avg acc \\ \hline
		Train from scratch & 3.53  & 81.32          & 36.08       & 47.0  & 3.35  & 34.26 \\
		Pretraining & 4.77 & 78.0    & 41.09  & 48.45 & 3.93  & 35.25 \\
		FOMAML (Finn et al. 2017)  & 9.8 & 77.8   & 42.27& \textbf{49.76} & 6.39   & 37.20 \\ \hline
		\textbf{AMS-FOMAML}  & \textbf{10.8} & \textbf{82.36} & \textbf{45.70} & 49.09 & \textbf{10.21} & \textbf{39.63} \\ \bottomrule
	\end{tabular}
	\caption{Results of speech classification in terms of accuracy(\%).}
	\label{tab:speech_classification}
% 	\vspace{-0.4em}
\end{table}

\begin{table}[]
	\centering
	\scriptsize
	
% 	\vspace{-0.5em}
	\begin{tabular}{l|ccc}
		\toprule 
		Method  & Mongolian & Swedish   & Turkish  \\ \hline
		CoVoST scratch~\cite{Wang2020CoVoSTAD} & 0.20 & 0.30 & 0.80 \\ \hline
		Train from scratch  & 0.27      & 0.22      & 0.85\\
		Pretraining   & 0.30      & 0.57      & 1.26 \\
		FOMAML (Finn et al. 2017) & 0.35      & 0.67      & 1.26\\ \hline
		\textbf{AMS-FOMAML} & \textbf{0.36} & \textbf{0.70} & \textbf{1.45}  \\ \bottomrule
	\end{tabular}
	\caption{Results of speech translation in terms of BLEU.}
	\label{tab:speech_translation}
% 	\vspace{-0.5em}
\end{table}

\noindent\textbf{AMS on speech translation.}
%We also evaluate our AMS on speech translation task. 
Here we consider translating other languages speech to   English.
% \noindent\textbf{Datasets.}
% CoVoST~\cite{Wang2020CoVoSTAD}, a multilingual speech-to-text translation (ST) corpus based on Common Voice corpus, has speech data in 11 languages and corresponding transcripts in English. 
We select 8 source languages of 10 hours and 3 target languages less than 10 hours from CoVoST~\cite{Wang2020CoVoSTAD}, a multilingual speech translation (ST) corpus. % Aiming to explore the relationship between these languages and English, we limit the quantity of each source language data to 10 hours. Besides, in order to prove the effectiveness of our methods, we also use the remaining 3 language includings Turkish (Tr), Swedish (Sv) and Mongolian (Mn) as target tasks, for the reason that their duration of speech data is less than 10 hours.
% \noindent\textbf{Experimental setup.}
For simplicity, we use the same model architecture and data preprocessing procedure as ASR in Sec.~\ref{low_asr_setup}, which can achieve the same performance as the model used  in CoVoST.
%and compare our method with 3 other methods mentioned in speech classification. %We compare our method with 3 methods, which are directly training on target tasks, combining speeches from all source task to get the initialization parameters of model through pretraining and FOMAML respectively, and use the target tasks to finetune.
% \noindent\textbf{Results.}
Table~\ref{tab:speech_translation} shows the result of case-insensitive tokenized BLEU~\cite{Papineni2002BleuAM} using sacreBLEU~\cite{Post2018ACF}. By comparison,  our method outperforms all baselines, including CoVoST, and achieves state-of-the-art performance in all the three target languages. %Furthermore, although our model architecture is different with the one used in CoVoST~\cite{Wang2020CoVoSTAD}, but the result of training from scratch is not too far apart. And compared with the method which combined 87 hours speech from French (Fr) with target language to train the model, our method outperform it in Mongolian and Swedish, except Turkish. Maybe the Fr data is more similar to the Turkish than other languages. 

% \noindent\textbf{Analysis of language difficulty for English.}
% 我们控制了speech translation每个source task的数量均等，用以探究其sample task time与speech translation source task种类的关系。
% 因为我们选择的speech translation task是从一种语言的语音翻译成英语的文本，因此sample task time次数越多，说明这种语言翻译成英语越困难，也说明这种语言与英语之间的相似度可能越小。
% 我们通过~\cite{Senel2018GeneratingSS,languagedifficulty}探究英语与其他语言间的semantic similarity和英语speaker学习其他语言的难度，并据此以英语为中心划分了四个虚线圈和实验中其他语言的位置，越外面的虚线圈代表着语言距离英语的距离越远。
% 同时通过实验数据，将sample times用该语言的圆形面积代替。
% 我们发现，sample次数最多的task，chinese，正是距离英语最远，对于英语speaker来说最难的task。而sample次数最少的task,dutch，也是距离英语很近的task。
% 图像整体也呈现出距离近的sample times少，距离远的sample times多的情形，除了Persian和french，但也有些研究~\cite{Georgi2010ComparingLS}认为persian与英语的距离是比较近的，同时法语由于Challenging Pronunciations, complex Spelling,  Strangely grammar， 其实也是很难的一门语言。
% 因此我们的方法确实还考虑到了task种类难度的imbalance，在每个step时倾向于sample出更加难的task for effective learning。

\noindent\textbf{Analysis of difficulty imbalance.}
\label{anlysis:difficult_imbalance}
We limit the quantity of each source language to 10 hours to analyze the task difficulty in our ST experiments. In Figure~\ref{fig:relation}, we use the semantic similarity between English and other source languages~\cite{Senel2018GeneratingSS} as well as the language learning difficulty for English speakers~\cite{languagedifficulty} as references to measure the distances, where the farther the language is, the more difficult is to translate it into English. It can be observed a trend that the sample times increase with the distances. For example, Chinese and Dutch are the farthest language with the highest sampling times and the closest language with the fewest sample times, respectively. This indicates that our method can automatically sample tasks according to the task difficulty to alleviate the imbalance from different language difficulties.

\section{Conclusion}
In this work, to tackle the task-imbalance problem caused by language tasks difficulties and quantities, we develop a novel Adversarial Meta Sampling framework to adaptively sample language tasks for learning a better model initialization for target low-resource languages. It can well handle the challenging multilingual low-resource ASR in real world.  Extensive experimental results  validate that our method  effectively improves the few-shot learning ability of  both meta-learning and transfer learning and also shows its great generalization capacity in other low-resource speech tasks.

\clearpage
\section*{Acknowledgements}
This work was supported in part by National Natural Science Foundation of China (NSFC) under Grant No.U19A2073 and No.61976233, Guangdong Province Basic and Applied Basic Research (Regional Joint Fund-Key) Grant No.2019B1515120039, Nature Science Foundation of Shenzhen Under Grant No. 2019191361, Zhijiang Lab’s Open Fund (No. 2020AA3AB14) and CSIG Young Fellow Support Fund.

\bibliography{aaai_2020}

\begin{thebibliography}{47}
\providecommand{\natexlab}[1]{#1}
\providecommand{\url}[1]{\texttt{#1}}
\providecommand{\urlprefix}{URL }
\expandafter\ifx\csname urlstyle\endcsname\relax
  \providecommand{\doi}[1]{doi:\discretionary{}{}{}#1}\else
  \providecommand{\doi}{doi:\discretionary{}{}{}\begingroup
  \urlstyle{rm}\Url}\fi

\bibitem[{Adams et~al.(2019)Adams, Wiesner, Watanabe, and
  Yarowsky}]{Adams2019MassivelyMA}
Adams, O.; Wiesner, M.; Watanabe, S.; and Yarowsky, D. 2019.
\newblock Massively Multilingual Adversarial Speech Recognition.
\newblock \emph{NAACL-HLT} 96--108.

\bibitem[{Arandjelovi{\'c} et~al.(2016)Arandjelovi{\'c}, Gron{\'a}t, Torii,
  Pajdla, and Sivic}]{Arandjelovi2016NetVLADCA}
Arandjelovi{\'c}, R.; Gron{\'a}t, P.; Torii, A.; Pajdla, T.; and Sivic, J.
  2016.
\newblock NetVLAD: CNN Architecture for Weakly Supervised Place Recognition.
\newblock \emph{2016 IEEE Conference on Computer Vision and Pattern Recognition
  (CVPR)} 5297--5307.

\bibitem[{Ardila et~al.(2020)Ardila, Branson, Davis, Henretty, Kohler, Meyer,
  Morais, Saunders, Tyers, and Weber}]{commonV}
Ardila, R.; Branson, M.; Davis, K.; Henretty, M.; Kohler, M.; Meyer, J.;
  Morais, R.; Saunders, L.; Tyers, F.~M.; and Weber, G. 2020.
\newblock Common Voice: A Massively-Multilingual Speech Corpus.
\newblock In \emph{LREC}.

\bibitem[{Chan et~al.(2016)Chan, Jaitly, Le, and Vinyals}]{Chan2016ListenAA}
Chan, W.; Jaitly, N.; Le, Q.~V.; and Vinyals, O. 2016.
\newblock Listen, attend and spell: A neural network for large vocabulary
  conversational speech recognition.
\newblock \emph{2016 IEEE International Conference on Acoustics, Speech and
  Signal Processing (ICASSP)} 4960--4964.

\bibitem[{Chen et~al.(2020)Chen, Hsu, Lee, and yi~Lee}]{Chen2020DARTSASRDA}
Chen, Y.-C.; Hsu, J.-Y.; Lee, C.-W.; and yi~Lee, H. 2020.
\newblock DARTS-ASR: Differentiable Architecture Search for Multilingual Speech
  Recognition and Adaptation.
\newblock In \emph{INTERSPEECH}.

\bibitem[{Cho et~al.(2018)Cho, Baskar, Li, Wiesner, Mallidi, Yalta,
  Karafi{\'a}t, Watanabe, and Hori}]{Cho2018MultilingualSS}
Cho, J.; Baskar, M.~K.; Li, R.; Wiesner, M.; Mallidi, S.~H.; Yalta, N.;
  Karafi{\'a}t, M.; Watanabe, S.; and Hori, T. 2018.
\newblock Multilingual Sequence-to-Sequence Speech Recognition: Architecture,
  Transfer Learning, and Language Modeling.
\newblock \emph{2018 IEEE Spoken Language Technology Workshop (SLT)} 521--527.

\bibitem[{Chorowski et~al.(2015)Chorowski, Bahdanau, Serdyuk, Cho, and
  Bengio}]{Chorowski2015AttentionBasedMF}
Chorowski, J.; Bahdanau, D.; Serdyuk, D.; Cho, K.; and Bengio, Y. 2015.
\newblock Attention-Based Models for Speech Recognition.
\newblock In \emph{NIPS}.

\bibitem[{Chung and Glass(2020)}]{Chung2019GenerativePF}
Chung, Y.-A.; and Glass, J. 2020.
\newblock Generative Pre-Training for Speech with Autoregressive Predictive
  Coding.
\newblock \emph{2020 IEEE International Conference on Acoustics, Speech and
  Signal Processing (ICASSP)} .

\bibitem[{Dalmia et~al.(2018)Dalmia, Sanabria, Metze, and
  Black}]{Dalmia2018SequenceBasedML}
Dalmia, S.; Sanabria, R.; Metze, F.; and Black, A.~W. 2018.
\newblock Sequence-Based Multi-Lingual Low Resource Speech Recognition.
\newblock \emph{2018 IEEE International Conference on Acoustics, Speech and
  Signal Processing (ICASSP)} 4909--4913.

\bibitem[{Dou, Yu, and Anastasopoulos(2019)}]{Dou2019InvestigatingMA}
Dou, Z.-Y.; Yu, K.; and Anastasopoulos, A. 2019.
\newblock Investigating Meta-Learning Algorithms for Low-Resource Natural
  Language Understanding Tasks.
\newblock In \emph{EMNLP/IJCNLP}.

\bibitem[{Finn, Abbeel, and Levine(2017)}]{Finn2017ModelAgnosticMF}
Finn, C.; Abbeel, P.; and Levine, S. 2017.
\newblock Model-Agnostic Meta-Learning for Fast Adaptation of Deep Networks.
\newblock In \emph{ICML}.

\bibitem[{FSI.(2007)}]{languagedifficulty}
FSI. 2007.
\newblock Language Learning Difficulty for English Speakers.
\newblock
  \url{https://en.wikibooks.org/wiki/Wikibooks:Language_Learning_Difficulty_for_English_Speakers}.

\bibitem[{Gales et~al.(2014)Gales, Knill, Ragni, and Rath}]{Gales2014SpeechRA}
Gales, M.; Knill, K.; Ragni, A.; and Rath, S.~P. 2014.
\newblock Speech recognition and keyword spotting for low-resource languages:
  Babel project research at CUED.
\newblock In \emph{SLTU}.

\bibitem[{Ganin et~al.(2016)Ganin, Ustinova, Ajakan, Germain, Larochelle,
  Laviolette, Marchand, and Lempitsky}]{Ganin2016DomainAdversarialTO}
Ganin, Y.; Ustinova, E.; Ajakan, H.; Germain, P.; Larochelle, H.; Laviolette,
  F.; Marchand, M.; and Lempitsky, V.~S. 2016.
\newblock Domain-Adversarial Training of Neural Networks.
\newblock \emph{Journal of Machine Learning Research} vol. 17, no. 1, pp.
  2096–2030.

\bibitem[{Graves et~al.(2006)Graves, Fern{\'a}ndez, Gomez, and
  Schmidhuber}]{Graves2006ConnectionistTC}
Graves, A.; Fern{\'a}ndez, S.; Gomez, F.~J.; and Schmidhuber, J. 2006.
\newblock Connectionist temporal classification: labelling unsegmented sequence
  data with recurrent neural networks.
\newblock In \emph{ICML '06}.

\bibitem[{Graves and Jaitly(2014)}]{Graves2014TowardsES}
Graves, A.; and Jaitly, N. 2014.
\newblock Towards End-To-End Speech Recognition with Recurrent Neural Networks.
\newblock In \emph{ICML}.

\bibitem[{Graves, Jaitly, and rahman Mohamed(2013)}]{Graves2013HybridSR}
Graves, A.; Jaitly, N.; and rahman Mohamed, A. 2013.
\newblock Hybrid speech recognition with Deep Bidirectional LSTM.
\newblock \emph{2013 IEEE Workshop on Automatic Speech Recognition and
  Understanding} 273--278.

\bibitem[{Hochreiter and Schmidhuber(1997)}]{Hochreiter1997LongSM}
Hochreiter, S.; and Schmidhuber, J. 1997.
\newblock Long Short-Term Memory.
\newblock \emph{Neural Computation} 9: 1735--1780.

\bibitem[{Hori et~al.(2017)Hori, Watanabe, Zhang, and
  Chan}]{Hori2017AdvancesIJ}
Hori, T.; Watanabe, S.; Zhang, Y.~L.; and Chan, W. 2017.
\newblock Advances in Joint CTC-Attention based End-to-End Speech Recognition
  with a Deep CNN Encoder and RNN-LM.
\newblock In \emph{INTERSPEECH}.

\bibitem[{Hsu, Chen, and yi~Lee(2020)}]{Hsu2019MetaLF}
Hsu, J.-Y.; Chen, Y.-J.; and yi~Lee, H. 2020.
\newblock Meta Learning for End-to-End Low-Resource Speech Recognition.
\newblock \emph{2020 IEEE International Conference on Acoustics, Speech and
  Signal Processing (ICASSP)} .

\bibitem[{Hu et~al.(2019)Hu, Bruguier, Sainath, Prabhavalkar, and
  Pundak}]{Hu2019}
Hu, K.; Bruguier, A.; Sainath, T.~N.; Prabhavalkar, R.; and Pundak, G. 2019.
\newblock {Phoneme-Based Contextualization for Cross-Lingual Speech Recognition
  in End-to-End Models}.
\newblock In \emph{Proc. Interspeech 2019}, 2155--2159.

\bibitem[{Kahn, Lee, and Hannun(2020)}]{Kahn2019SelfTrainingFE}
Kahn, J.; Lee, A.; and Hannun, A. 2020.
\newblock Self-Training for End-to-End Speech Recognition.
\newblock \emph{2020 IEEE International Conference on Acoustics, Speech and
  Signal Processing (ICASSP)} .

\bibitem[{Kim, Hori, and Watanabe(2017)}]{Kim2017JointCB}
Kim, S.; Hori, T.; and Watanabe, S. 2017.
\newblock Joint CTC-attention based end-to-end speech recognition using
  multi-task learning.
\newblock \emph{2017 IEEE International Conference on Acoustics, Speech and
  Signal Processing (ICASSP)} 4835--4839.

\bibitem[{Kunze et~al.(2017)Kunze, Kirsch, Kurenkov, Krug, Johannsmeier, and
  Stober}]{Kunze2017TransferLF}
Kunze, J.; Kirsch, L.; Kurenkov, I.; Krug, A.; Johannsmeier, J.; and Stober, S.
  2017.
\newblock Transfer Learning for Speech Recognition on a Budget.
\newblock In \emph{Rep4NLP,ACL}.

\bibitem[{Li et~al.(2019)Li, Sainath, Pang, and Wu}]{Li2019SemisupervisedTF}
Li, B.; Sainath, T.~N.; Pang, R.; and Wu, Z. 2019.
\newblock Semi-supervised Training for End-to-end Models via Weak Distillation.
\newblock \emph{IEEE International Conference on Acoustics, Speech and Signal
  Processing (ICASSP)} 2837--2841.

\bibitem[{Nichol, Achiam, and Schulman(2018)}]{Nichol2018OnFM}
Nichol, A.; Achiam, J.; and Schulman, J. 2018.
\newblock On First-Order Meta-Learning Algorithms.
\newblock \emph{ArXiv} abs/1803.02999.

\bibitem[{Papineni et~al.(2002)Papineni, Roukos, Ward, and
  Zhu}]{Papineni2002BleuAM}
Papineni, K.; Roukos, S.; Ward, T.; and Zhu, W.-J. 2002.
\newblock Bleu: a Method for Automatic Evaluation of Machine Translation.
\newblock In \emph{ACL}.

\bibitem[{Post(2018)}]{Post2018ACF}
Post, M. 2018.
\newblock A Call for Clarity in Reporting BLEU Scores.
\newblock In \emph{WMT}.

\bibitem[{Pratap et~al.(2019)Pratap, Hannun, Xu, Cai, Kahn, Synnaeve,
  Liptchinsky, and Collobert}]{Pratap2019Wav2LetterAF}
Pratap, V.; Hannun, A.; Xu, Q.; Cai, J.; Kahn, J.; Synnaeve, G.; Liptchinsky,
  V.; and Collobert, R. 2019.
\newblock Wav2Letter++: A Fast Open-source Speech Recognition System.
\newblock \emph{IEEE International Conference on Acoustics, Speech and Signal
  Processing (ICASSP)} 6460--6464.

\bibitem[{Sandler et~al.(2018)Sandler, Howard, Zhu, Zhmoginov, and
  Chen}]{Sandler2018MobileNetV2IR}
Sandler, M.; Howard, A.~G.; Zhu, M.; Zhmoginov, A.; and Chen, L.-C. 2018.
\newblock MobileNetV2: Inverted Residuals and Linear Bottlenecks.
\newblock \emph{2018 IEEE/CVF Conference on Computer Vision and Pattern
  Recognition} 4510--4520.

\bibitem[{Schneider et~al.(2019)Schneider, Baevski, Collobert, and
  Auli}]{Schneider2019wav2vecUP}
Schneider, S.; Baevski, A.; Collobert, R.; and Auli, M. 2019.
\newblock wav2vec: Unsupervised Pre-training for Speech Recognition.
\newblock In \emph{INTERSPEECH 2019}.

\bibitem[{Senel et~al.(2018)Senel, Utlu, Y{\"u}cesoy, Koc, and
  Çukur}]{Senel2018GeneratingSS}
Senel, L.~K.; Utlu, I.; Y{\"u}cesoy, V.; Koc, A.; and Çukur, T. 2018.
\newblock Generating Semantic Similarity Atlas for Natural Languages.
\newblock \emph{2018 IEEE Spoken Language Technology Workshop (SLT)} 795--799.

\bibitem[{Shinohara(2016)}]{Shinohara2016AdversarialML}
Shinohara, Y. 2016.
\newblock Adversarial Multi-Task Learning of Deep Neural Networks for Robust
  Speech Recognition.
\newblock In \emph{INTERSPEECH}.

\bibitem[{Simonyan and Zisserman(2015)}]{Simonyan2015VeryDC}
Simonyan, K.; and Zisserman, A. 2015.
\newblock Very Deep Convolutional Networks for Large-Scale Image Recognition.
\newblock \emph{CoRR} abs/1409.1556.

\bibitem[{Sun et~al.(2018{\natexlab{a}})Sun, Liu, Chua, and
  Schiele}]{Sun2018MetaTransferLF}
Sun, Q.; Liu, Y.; Chua, T.-S.; and Schiele, B. 2018{\natexlab{a}}.
\newblock Meta-Transfer Learning for Few-Shot Learning.
\newblock \emph{2019 IEEE/CVF Conference on Computer Vision and Pattern
  Recognition (CVPR)} 403--412.

\bibitem[{Sun et~al.(2018{\natexlab{b}})Sun, Yeh, Hwang, Ostendorf, and
  Xie}]{Sun2018DomainAT}
Sun, S.; Yeh, C.-F.; Hwang, M.-Y.; Ostendorf, M.; and Xie, L.
  2018{\natexlab{b}}.
\newblock Domain Adversarial Training for Accented Speech Recognition.
\newblock \emph{2018 IEEE International Conference on Acoustics, Speech and
  Signal Processing (ICASSP)} 4854--4858.

\bibitem[{Tong, Garner, and Bourlard(2017)}]{Tong2017MultilingualTA}
Tong, S.; Garner, P.~N.; and Bourlard, H. 2017.
\newblock Multilingual Training and Cross-lingual Adaptation on CTC-based
  Acoustic Model.
\newblock \emph{ArXiv} abs/1711.10025.

\bibitem[{Toshniwal et~al.(2018)Toshniwal, Sainath, Weiss, Li, Moreno,
  Weinstein, and Rao}]{Toshniwal2018MultilingualSR}
Toshniwal, S.; Sainath, T.~N.; Weiss, R.~J.; Li, B.; Moreno, P.~J.; Weinstein,
  E.; and Rao, K. 2018.
\newblock Multilingual Speech Recognition with a Single End-to-End Model.
\newblock \emph{2018 IEEE International Conference on Acoustics, Speech and
  Signal Processing (ICASSP)} 4904--4908.

\bibitem[{Waibel et~al.(2000)Waibel, Soltau, Schultz, Schaaf, and
  Metze}]{Waibel2000}
Waibel, A.; Soltau, H.; Schultz, T.; Schaaf, T.; and Metze, F. 2000.
\newblock \emph{Multilingual Speech Recognition}, 33--45.
\newblock Springer Berlin Heidelberg.

\bibitem[{Wang et~al.(2020)Wang, Pino, Wu, and Gu}]{Wang2020CoVoSTAD}
Wang, C.; Pino, J.; Wu, A.; and Gu, J. 2020.
\newblock CoVoST: A Diverse Multilingual Speech-To-Text Translation Corpus.
\newblock \emph{ArXiv} abs/2002.01320.

\bibitem[{Wang, Tsvetkov, and Neubig(2020)}]{wang-etal-2020-balancing}
Wang, X.; Tsvetkov, Y.; and Neubig, G. 2020.
\newblock Balancing Training for Multilingual Neural Machine Translation.
\newblock In \emph{Proceedings of the 58th Annual Meeting of the Association
  for Computational Linguistics}, 8526--8537.

\bibitem[{Watanabe, Hori, and Hershey(2017)}]{Watanabe2017LanguageIE}
Watanabe, S.; Hori, T.; and Hershey, J.~R. 2017.
\newblock Language independent end-to-end architecture for joint language
  identification and speech recognition.
\newblock \emph{IEEE Automatic Speech Recognition and Understanding Workshop
  (ASRU)} 265--271.

\bibitem[{Williams(1992)}]{Williams1992SimpleSG}
Williams, R.~J. 1992.
\newblock Simple Statistical Gradient-Following Algorithms for Connectionist
  Reinforcement Learning.
\newblock \emph{Machine Learning} 8: 229--256.

\bibitem[{Winata et~al.(2020)Winata, Cahyawijaya, Lin, Liu, Xu, and
  Fung}]{Winata2020MetaTransferLF}
Winata, G.~I.; Cahyawijaya, S.; Lin, Z.; Liu, Z.; Xu, P.; and Fung, P. 2020.
\newblock Meta-Transfer Learning for Code-Switched Speech Recognition.
\newblock In \emph{ACL}.

\bibitem[{Yi et~al.(2018)Yi, Tao, Wen, and Bai}]{Yi2018AdversarialMT}
Yi, J.; Tao, J.; Wen, Z.; and Bai, Y. 2018.
\newblock Adversarial Multilingual Training for Low-Resource Speech
  Recognition.
\newblock \emph{2018 IEEE International Conference on Acoustics, Speech and
  Signal Processing (ICASSP)} 4899--4903.

\bibitem[{Zhou et~al.(2019)Zhou, Yuan, Xu, Yan, and Feng}]{Zhou2019EfficientML}
Zhou, P.; Yuan, X.; Xu, H.; Yan, S.; and Feng, J. 2019.
\newblock Efficient Meta Learning via Minibatch Proximal Update.
\newblock In \emph{NeurIPS}.

\bibitem[{Zhou et~al.(2020)Zhou, Zou, Yuan, Feng, Xiong, and
  Hoi}]{Zhou2020MAML}
Zhou, P.; Zou, Y.; Yuan, X.; Feng, J.; Xiong, C.; and Hoi, S.~C. 2020.
\newblock Task Similarity Aware Meta Learning: Theory-inspired Improvement on
  MAML.
\newblock In \emph{4th Workshop on Meta-Learning at NeurIPS}.

\end{thebibliography}

\end{document}